\documentclass[a4paper,fleqn]{cas-dc}
\pdfoutput=1
\usepackage[numbers]{natbib}

\usepackage{stmaryrd}
\usepackage{epsfig}
\usepackage{graphicx}
\usepackage{bm}

\usepackage{amsfonts}
\DeclareGraphicsExtensions{.eps}
\usepackage{algorithmic}

\usepackage{multirow}
\usepackage{hhline}
\usepackage{epstopdf}

\usepackage{dsfont}
\usepackage{url}
\usepackage{hyperref}
\usepackage{diagbox}
\usepackage{threeparttable}
\usepackage{pdfpages}
\usepackage{fancyref}
\usepackage{subfigure}
\usepackage{ulem}
\usepackage{xspace}

\def\tsc#1{\csdef{#1}{\textsc{\lowercase{#1}}\xspace}}
\tsc{WGM}
\tsc{QE}
\tsc{EP}
\tsc{PMS}
\tsc{BEC}
\tsc{DE}

\begin{document}
\let\WriteBookmarks\relax
\def\floatpagepagefraction{1}
\def\textpagefraction{.001}

\shorttitle{arXiv}

\shortauthors{J. Shi et~al.}

\title [mode = title]{Distilled Low Rank Neural Radiance Field with Quantization for Light Field Compression}                      



%
\author{Jinglei Shi}[
        orcid=0000-0003-2926-0415
        ]


\ead{jinglei.shi@inria.fr}

\author{Christine Guillemot}[
        orcid=0000-0003-1604-967X
        ]

\ead{christine.guillemot@inria.fr}

\address{Institut National de Recherche en Informatique et en Automatique (INRIA), 263 Av. Général Leclerc, Rennes, 35042, France}

\begin{abstract}
We propose in this paper a Quantized Distilled Low-Rank Neural Radiance Field (QDLR-NeRF) representation for the task of light field compression. While existing compression methods encode the set of light field sub-aperture images, our proposed method learns an implicit scene representation in the form of a Neural Radiance Field (NeRF), which also enables view synthesis. To reduce its size, the model is first learned under a Low-Rank (LR) constraint using a Tensor Train (TT) decomposition within an Alternating Direction Method of Multipliers (ADMM) optimization framework. To further reduce the model's size, the components of the tensor train decomposition need to be quantized. However, simultaneously considering the optimization of the NeRF model with both the low-rank constraint and rate-constrained weight quantization is challenging. To address this difficulty, we introduce a network distillation operation that separates the low-rank approximation and the weight quantization during network training. The information from the initial LR-constrained NeRF (LR-NeRF) is distilled into a model of much smaller dimension (DLR-NeRF) based on the TT decomposition of the LR-NeRF. We then learn an optimized global codebook to quantize all TT components, producing the final QDLR-NeRF. Experimental results show that our proposed method yields better compression efficiency compared to state-of-the-art methods, and it additionally has the advantage of allowing the synthesis of any light field view with high quality.
\end{abstract}


\maketitle

\section{Introduction}
Light field imaging has garnered significant attention in recent years, driven by its immense potential for applications in computer vision and computational photography. Nevertheless, light fields inherently entail vast amounts of data, necessitating the development of compact representations and efficient compression algorithms. To address this challenge, numerous solutions have been devised for compressing light fields. Some solutions encode the input views using video compression techniques such as HEVC (High-Efficiency Video Coding), or variations of this standard \cite{Conti2016Bi, Ahmad2019}.
Light field views are rearranged into a pseudo video sequence (PVS) before being processed by the video codec. Generally, a predetermined scanning pattern, such as raster or spiral~\cite{Raster,amirpour2018high}, is employed to create the PVS.
Approaches based on low rank models \cite{FDL, INRIA_LF_dataset} or 4D steered Gaussian mixture of experts \cite{TMM1} have also been proposed for light field compression. 
One notable technique is the Multidimensional Light field Encoder (MuLE)~\cite{mule}, which has been incorporated into the JPEG Pleno Coding standard as a transform-based coding approach. MuLE leverages the 4D redundancy of light field images by segmenting them into 4D blocks and subsequently applying a 4D-DCT transform to exploit this redundancy \cite{4D-DCT}. Additionally, research has delved into the use of 6D transforms for plenoptic point cloud compression, as explored in \cite{krivokuca2021compression}.
In a related study, Liu et al. \cite{liu2019content} presented a prediction method that utilizes Gaussian Process Regression, applied to various classes of light field textures. Other approaches involve synthesizing the complete light field from a subset of input views, aided by estimated disparity maps \cite{tb, DIBR1, zhang2022light, huang2020low}.

Here, we present a novel approach centered on Neural Radiance Fields (NeRF), a concept introduced recently for scene representation and light field view synthesis \cite{mildenhall2020nerf}. NeRF operates as an implicit Multi-Layer Perceptron-based model, mapping 5D vectors—comprising 3D coordinates and 2D viewing directions—to opacity and color values. After being trained on a set of input images, NeRF will become capable of generating views observed from any desired perspectives by applying the volume rendering techniques. Various adaptations of NeRF have emerged later, aimed at reducing the number of required input views, as seen in \cite{Yu2021pixelNeRFNR}, or enhancing rendering efficiency, as demonstrated in \cite{Liu2020NeuralSV}.
A generalizable radiance field reconstruction is also proposed in \cite{MVSNeRF} which can reconstruct radiance fields from only three nearby input views via fast network inference, and which can be used for rendering scenes different from those on which the network has been trained.
An end-to-end framework, called NeRF--, is proposed in \cite{Wang2021NeRFNR} for training NeRF models without pre-computed camera parameters. 
The authors in \cite{Bemana2020xfields} generalize the usage of implicit neural representation to solve the problem of time, light and view interpolation. The authors in \cite{jeong2021selfcalibrating} consider more complex non-linear camera models and propose a new geometric loss function to jointly learn the geometry of the scene and the camera parameters. 

NeRF can thus be seen as a neural representation of a scene which is learned from a set of input views. While existing light field compression solutions usually consider encoding and transmitting a subset of light field views which are then used at the decoder to synthesize and render the entire light field, we consider an alternative approach which optimizes and compresses the NeRF on the sender side, so that the network itself is transmitted to the receiver. The light field compression problem is thus cast into a problem of neural network compression. However, unlike compressing networks used in image classification or recognition tasks \cite{han2016deep}, compressing a view synthesis network is more challenging, due to the fact that the light field reconstruction quality is sensitive to network weight changes. 
The model should not only be compact, but should also preserve as much as possible high reconstruction quality.

In this paper, we first consider a simplified NeRF model based on only one Multi-Layer Perception (MLP) for rendering instead of the coarse and fine MLP in \cite{mildenhall2020nerf}, which halves the number of parameters of the network. 
We optimize the network  under a low rank constraint. The learning problem is formulated as a tensor rank optimization problem, solved using the Alternating Direction Method of Multipliers (ADMM) iterative optimization method. 
This low rank constraint allows us, following principles of network distillation \cite{Hinton2015DistillingTK}, to transfer knowledge from this initial NeRF to a model of a much smaller dimension based on a Tensor Train (TT) decomposition.
In other words, the weights of NeRF are decomposed into TT components using a distillation network, to decrease the number of parameters to be encoded. The TT parameters are further quantized using an optimized codebook.
The knowledge transfer is made possible thanks to the low rank constrained optimization of the large NeRF model.

We evaluated the compression efficiency of our proposed Quantized Distilled Low Rank Neural Radiance Field (QDLR-NeRF) for light fields by comparing it against several benchmarks. Specifically, we assessed its performance against a HEVC-based method \cite{SullivanHEVC, HEVCstandard} and the Jpeg-pleno standard \cite{jpegpleno} in the 4D Prediction Mode. Moreover, we conducted comparative analyses with recent deep learning video compression techniques, including Hierarchical Learned Video Compression (HLVC) \cite{yang2020Learning}, Recurrent Learned Video Compression (RLVC) \cite{yang2021learning}, and OpenDVC \cite{yang2020opendvc}, which is based on Deep Video Compression (DVC) \cite{Lu2019DVC}. These video compression methods were applied to the sequence of light field views. Although motion estimation-based deep compression networks achieve impressive rate-distortion performance, their intricate architectures can pose training challenges and require pre-trained optical flow estimators. Our experimental results demonstrate that our approach outperforms recent deep video compression methods, even those trained on extensive datasets. Additionally, our method competes effectively with established HEVC video compression tools and Jpeg-Pleno, a dedicated compression tool designed explicitly for light fields.

This manuscript serves as both an extension and a complement to our shorter conference paper \cite{shi2023light}, providing in-depth technical details, elaborating on our design philosophy, and presenting comprehensive experimental results. In summary, along with \cite{shi2023light}, our work contributes to the field in the following ways:

\begin{itemize}
    \item A low rank constrained simplified NeRF for compact light field representation.
    \item A network distillation operation that transfers knowledge from the LR-NeRF to a more compact scene light field representation model DLR-NeRF via Tensor Train decomposition.
    \item An efficient quantization operation guided by a global codebook, which tremendously reduces the total size of the representation with a limited distortion. 
    \item A complete light field compression algorithm
    that integrates ADMM-based low rank optimization, network distillation and optimized quantization, giving superior performances when compared with
    light field reference compression methods.
\end{itemize}
Note that the proposed approach can be used beyond light field compression, in particular to reduce the memory footprint of NeRF, which can be critical for devices with low memory resources.

\begin{figure*}[!h]
    \centering
  \includegraphics[width=0.99\linewidth]{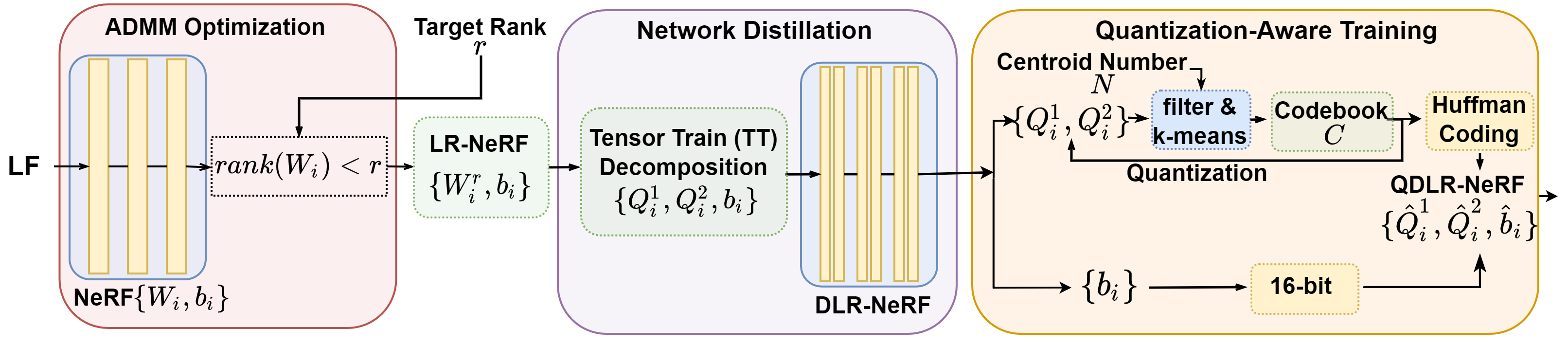}
  \caption{Overall workflow of our proposed QDLR-NeRF method. A simplified NeRF (formed by only one MLP) is first trained from the set of input views, under a low rank (LR) constraint. The information from this LR-NeRF is distilled to a model of a much smaller dimension DLR-NeRF via the Tensor Train decomposition. The weight distillation is used to transfer knowledge in parameters of the large LR-NeRF to a smaller DLR-NeRF while keeping high performance. A codebook $C$ with $N$ centroids is learned to quantize DLR-NeRF to obtain the final compact QDLR-NeRF.}
  \label{fig:workflow}
\end{figure*}

\section{Workflow and notations}
We consider a light field represented by a 4D function $L(x,y,u,v)$ \cite{levoy1996light}, with $(x,y)\in\llbracket 1,X \rrbracket \times \llbracket 1,Y \rrbracket$ 
and $(u,v)\in\llbracket 1,U \rrbracket \times \llbracket 1,V \rrbracket$ respectively its angular and spatial coordinates.

The overall workflow of our method is shown in Fig~\ref{fig:workflow}. The workflow is divided into three phases: 1) The low rank-constrained neural radiance field (LR-NeRF) optimization for scene reconstruction. 2) The network distillation for transferring knowledge from the LR-NeRF to a smaller model DLR-NeRF based on the Tensor Train decomposition. 3) The quantization of the TT components with an optimized codebook.  

In the first step, a simplified NeRF with one MLP having weight matrices $\bm{W}_{i}$ and bias vectors $\bm{b}_{i}$ in the layer indexed by $i$ is trained as an implicit scene representation, with the goal of having the best scene reconstruction given the set of input views. For the sake of simplicity, we abuse the term of `weight' to indicate both $\bm{W}_{i}$ and $\bm{b}_{i}$ in the rest of the paper. The obtained NeRF is finetuned under a low rank constraint by applying Alternating Direction Method of Multipliers (ADMM) optimization. The tensor of the finetuned weights of LR-NeRF $\{\bm{W}^{r}_{i},\bm{b}_{i}\}$ becomes low-rank after this step, where $r$ is the target rank. In the network distillation step, the weights of LR-NeRF are decomposed into a Tensor Train (TT) format as $\bm{W}^{r}_{i} = \bm{Q}^{1}_{i}\cdot \bm{Q}^{2}_{i}$, and we use $\{\bm{Q}^{1}_{i}, \bm{Q}^{2}_{i}, \bm{b}_{i}\}$ to initialize a distilled version of LR-NeRF noted DLR-NeRF. This step bonds the low rank optimization in previous step and the quantization in the next step. It can, on one hand, guarantee low rank property of $\bm{W}^{r}_{i}$, on the other hand, allow quantifying $\bm{Q}^{1}_{i}, \bm{Q}^{2}_{i}$ without degrading the final compression performance. The last quantization step aims at reducing the model size by quantifying the network weights. A global codebook $C$ with $N$ centroids is learned from $\bm{Q}^{1}_{i}, \bm{Q}^{2}_{i}$ to quantify the values within the interval [-1,1] of DLR-NeRF, from the first layer to the last layer gradually. Larger values outside the interval [-1,1] are quantized with a uniform $16$-bit quantizer. Thereafter, both sets of values inside and outside the interval [-1,1] are coded using Huffman coding for optimal bitrate. The biases of the network $\bm{b}_{i}$ are converted to 16 bits for further compacting the model size. We denote the quantized weights of QDLR-NeRF as $\{\bm{\hat{Q}}^{1}_{i},\bm{\hat{Q}}^{2}_{i},\bm{\hat{b}}_{i}\}$  

\section{Method}
\subsection{Simplified Neural Radiance Field}
In our method, we adopt NeRF to implicitly represent the scene, where 5D coordinates (location and view direction) of light rays $(x, y, z, \theta, \phi)$ are fed into an MLP to produce a color and volume density $(RGB,\sigma)$ for volumetric rendering. This procedure can be formulated as:
\begin{equation}
    (RGB,\sigma) = F_{\Theta}(x,y,z,\theta,\phi),
\end{equation}
with $\Theta = \{\bm{W}_{i},\bm{b}_{i}\}$ being weights and biases of the MLP. 

Let us note that the original version of NeRF uses two MLPs to respectively coarsely and densely sample rays  (i.e. performing a hierarchical volume sampling). The camera pose parameters are estimated from given views using the COLMAP structure-from-motion package \cite{schonberger2016structure}. We instead adopt one 
unique MLP for rendering, which halves the number of parameters of the network. Since COLMAP may struggle to estimate camera pose parameters, particularly in cases of narrow-baseline light fields, we adopt the idea of treating camera pose parameters as trainable variables, as seen in \cite{Wang2021NeRFNR}. 
Fortunately, when dealing with structured light fields, the necessary camera pose parameters are greatly simplified. Only the focal length $f$ and a parallax offset $\Delta$, proportional to the angular distance, are required for ray tracing in NeRF. 
In this step, the rendering error is used as the loss function to supervise the initialization of NeRF and the update of its parameters $\{\bm{W}_{i},\bm{b}_{i},f,\Delta \}$ as: 
\begin{equation}
\label{eq:loss}
    argmin_{\{\bm{W}_{i},\bm{b}_{i},f,\Delta\}}||c'-c_{gt}||^{2}_{2},
\end{equation}
where $c'$ is the rendered pixel value and $c_{gt}$ is ground-truth pixel value.

\subsection{Low Rank NeRF approximation with ADMM}
Although the initial NeRF \cite{mildenhall2020nerf} gives a good scene reconstruction and could be used for light field compression, its model size makes transmitting such a network less efficient than transmitting the compressed light field views. We therefore further reduce the model size by applying a TT decomposition \cite{oseledets2011tensor} to the learned NeRF. The network weights $\bm{W}_{i}\in \mathbb{R}^{n_{1}\times n_{2}}, \forall i$ are decomposed into the TT format as
\begin{equation}
\label{eq:TT}
    \bm{W}_{i} = \bm{Q}^{1}_{i}\cdot \bm{Q}^{2}_{i},
\end{equation}
where $\bm{Q}^{1}_{i}\in \mathbb{R}^{1\times n_{1}\times r}$ and $\bm{Q}^{2}_{i}\in \mathbb{R}^{r\times n_{2}\times 1}$ are the TT cores, $r$ is the target rank controlling the compression ratio, $n_1$ and $n_2$ represent respectively the input and output channel numbers of the fully connected layer.
Straightforwardly decomposing the full-rank tensor $\bm{W}_{i}$ into a low-rank TT format inevitably causes a large approximation error, which  degrades the light field reconstruction quality and the compression efficiency. To guarantee the low rank property of the tensor $\bm{W}_{i}$ without performance degradation, the authors in \cite{yin2021towards} formulate the model compression as a tensor rank optimization problem, and leverage Alternating Direction Method of Multipliers(ADMM) to solve this optimization problem in an iterative manner. In this section, we adopt this ADMM-based low rank approximation to finetune the learned NeRF. This optimization can be formulated as
\begin{gather}
    argmin_{\{\bm{W}_{i},\bm{b}_{i}\}} ||c'-c_{gt}||^{2}_{2}\\
    s.t. \ \ rank(\bm{W}_{i}) < r,
\end{gather}
where $r$ is the desired rank of $\bm{W}_{i}$. 

This non-convex optimization with constraints can be solved using the ADMM optimization method, after introducing an auxiliary variable \bm{$Z_{i}$} and an indicator function $g(\cdot)$, as
\begin{equation}
\label{indicator_function}
    g(\bm{W}_{i}) = \!
    \begin{cases}
    0, rank(\bm{W}_{i}) < r \\
    +\infty, otherwise.
    \end{cases}
\end{equation}
The original optimization problem can then be rewritten as:
\begin{gather}
    argmin_{\{\bm{W}_{i},\bm{Z}_{i}\}} ||c'-c_{gt}||^{2}_{2} +g(\bm{Z}_{i})\\
    s.t. \ \  \bm{W}_{i} = \bm{Z}_{i}  
\end{gather}
and the corresponding augmented Lagrangian in the scaled dual form of such an optimization problem is defined as 
\begin{equation}
    \mathcal{L}(\bm{W}_{i},\bm{Z}_{i},\bm{U}_{i}) = l(\bm{W}_{i})+g(\bm{Z}_{i})+\frac{\rho}{2}||\bm{W}_{i}-\bm{Z}_{i}+\bm{U}_{i}||^2_{F}+\frac{\rho}{2}||\bm{U}_{i}||^{2}_{F},
\end{equation}
where $\rho$ is a positive penalty parameter, $\bm{U}_{i}$ is the dual multiplier, and $l = ||c'-c_{gt}||^2_{2}$ is the reconstruction error. The optimization of this Lagrangian term can be performed in an alternative way, as
\begin{align}
    \label{w_problem}
    \bm{W}_{i}^{t+1} &= argmin_{\{\bm{W}_{i}\}} \mathcal{L}(\bm{W}_{i}^{t},\bm{Z}_{i}^{t},\bm{U}_{i}^{t}),\\
    \label{z_problem}
    \bm{Z}_{i}^{t+1} &= argmin_{\{\bm{Z}_{i}\}}\mathcal{L}(\bm{W}_{i}^{t+1},\bm{Z}_{i}^{t},\bm{U}_{i}^{t}),\\
    \bm{U}_{i}^{t+1} &= \bm{U}_{i}^{t} + \bm{W}_{i}^{t+1} - \bm{Z}_{i}^{t+1}
\end{align}
Eq.\ref{w_problem} corresponding to the updates of $\bm{W}_{i}$ can be solved using gradient descent. Eq.\ref{z_problem} can be solved according to \cite{boyd2011distributed} as
\begin{equation}
    \bm{Z}_{i}^{t+1} = \Pi_{r}(\bm{W}^{t+1}_{i}+\bm{U}^{t}_{i}),
\end{equation}
here, $\Pi_{r}$ represents the operation that decomposes the matrix into the TT format, truncated to the desired rank $r$. One can refer to \cite{yin2021towards} for more details on the ADMM procedure.

After this ADMM-based optimization procedure, the weights of the network $\bm{W}^{r}_{i}$ are low rank, hence can be decomposed into the TT format as in Eq.~\ref{eq:TT}, with a small approximation error. For sake of simplicity, in the following, we neglect the first dimension of $\bm{Q}^{1}_{i}$ and the last dimension of $\bm{Q}^{2}_{i}$, hence $\bm{Q}^{1}_{i}$ and $\bm{Q}^{2}_{i}$ are respectively with size $n_{1}\times r$ and $r\times n_{2}$. As the vector of bias parameters of each layer $\bm{b}_{i}$ is of rank 1, they are not considered in this optimization. In summary, after this low-rank optimization step, the parameters of the layers $\{\bm{W}^{r}_{i},\bm{b}_{i}\}$ are approximated by their TT components $\{\bm{Q}^{1}_{i},\bm{Q}^{2}_{i},\bm{b}_{i}\}$ for the following processing steps.

\begin{figure}[!t]
  \centering 
  \subfigure[Weight distribution before applying rate-constrained quantization.]{ 
    \label{fig:weight_distribution_before}
    \includegraphics[width=0.8\linewidth]{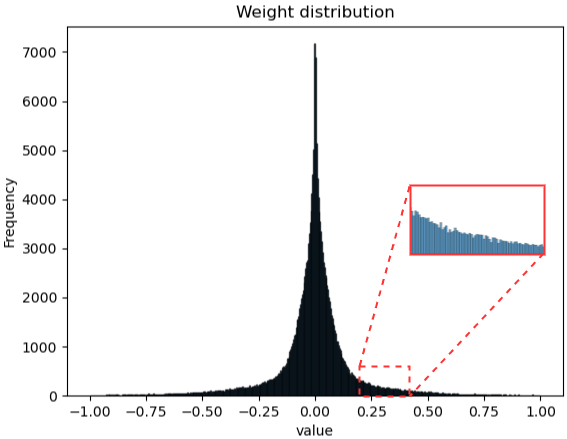}}\\
    \subfigure[Weight distribution after adopting rate-constrained quantization, with the number of centroid $N=256$.]{
    \label{fig:weight_distribution_after}
    \includegraphics[width=0.8\linewidth]{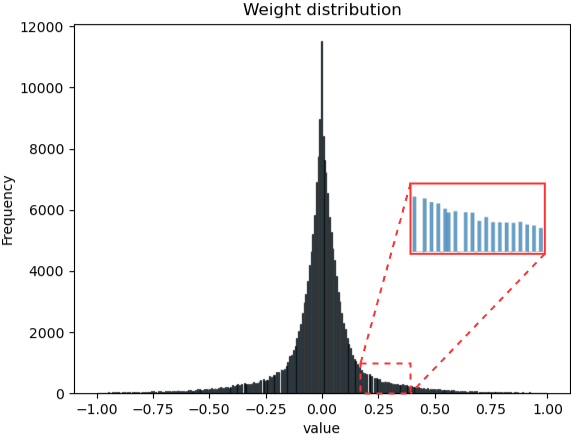}} 
    \vspace{-0.2cm}
  \caption{The weight distribution of coefficients of $\{\bm{Q}^{1}_{i},\bm{Q}^{2}_{i}\}$ before\&after applying rate-constrained quantization, we take light field scene `\textit{sideboard}' as an example, the target rank $r=40$ and number of centroid for quantization is $N=256$.} 
  \label{fig:weight_distribution}
\end{figure}

\begin{figure}[t]
    \centering
    \setlength{
    \tabcolsep}{1pt}
    \centering
    \begin{tabular}{cc}
\includegraphics[width=0.48\linewidth]{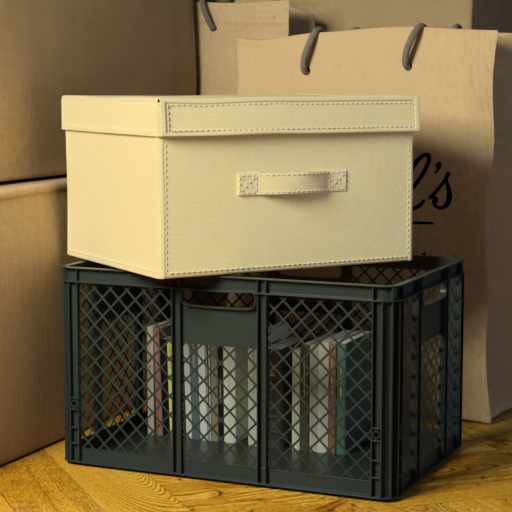} &
\includegraphics[width=0.48\linewidth]{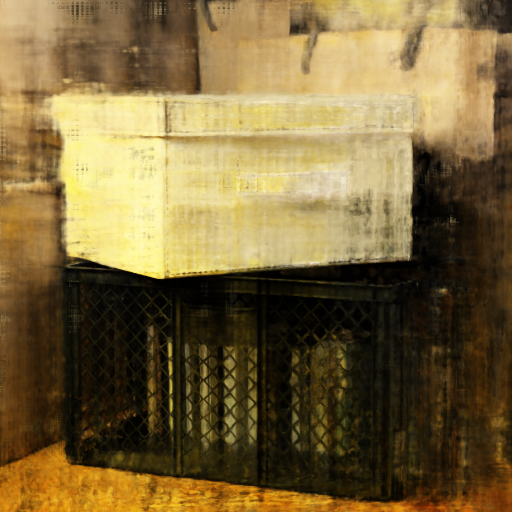} \\
\end{tabular}
\caption{Reconstructed views (`\textit{boxes}(5,5)') using network weights quantized in uniform 16 bit, without (left) and with (right) truncation by the interval [-1,1]. }
\label{fig:anchor_basic}
\end{figure}

\begin{figure}
    \centering
  \includegraphics[width=0.85\linewidth]{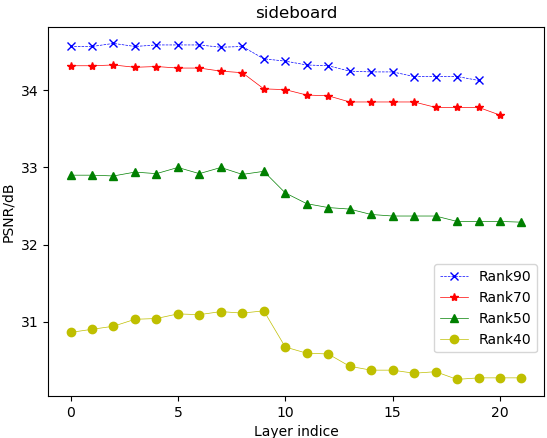}
  \caption{PSNR evolution after quantizing each layer. We take the light field scene `\textit{sideboard}' with diffrent ranks $r=\{ 40,50,70,90\}$ as an example.}
  \label{fig:quantization}
\end{figure}

\subsection{Rate-Constrained Quantization}
\label{sec:quantization}
\subsubsection{Quantization strategies}

To further compact the learned model, we employ different quantization strategies for the parameters in ${\bm{Q}^{1}_{i}, \bm{Q}^{2}_{i}}$ and ${\bm{b}_{i}}$. Given that the number of parameters in the bias vectors ${\bm{b}_{i}}$ is significantly smaller than that in the matrices ${\bm{Q}^{1}_{i}, \bm{Q}^{2}_{i}}$, we opt for a simple uniform 16-bit quantization (half precision in PyTorch) for the parameters in ${\bm{b}_{i}}$. However, for the parameters in ${\bm{Q}^{1}_{i}, \bm{Q}^{2}_{i}}$, we utilize a training-aware non-uniform quantization to minimize the required bit number.

The quantization strategy for $\{\bm{Q}^{1}_{i}, \bm{Q}^{2}_{i}\}$ is based on the characteristics of their parameter distribution.
Fig.~\ref{fig:weight_distribution_before} shows the distribution of parameters of $\{\bm{Q}^{1}_{i}, \bm{Q}^{2}_{i}\}$, where one can observe that most of the parameters are within the interval $[-1,1]$, while quite few (less than 0.5\% of the total number) parameters are coarsely located outside the interval [-1,1]. Although these parameters occupy only a small portion of the total network parameters, they are essential for a good light field reconstruction. As shown in Fig.~\ref{fig:anchor_basic}, the truncation of parameters outside the interval [-1, 1] causes severe image degradation, hence these parameters are of great importance and should retain in high precision. One the other hand, considering the small portion of these parameters, we carry out 16-bit uniform quantization (half precision in Pytorch) on them, which barely increases the final average bitrate.

The quantization strategy for ${\bm{Q}^{1}_{i}, \bm{Q}^{2}_{i}}$ is derived from the characteristics of their parameter distribution. In Fig.~\ref{fig:weight_distribution_before}, we can observe that the majority of these parameters fall within the interval $[-1,1]$, while a small fraction (less than 0.5\% of the total) are located outside this range. Despite occupying only a minor portion of the total network parameters, these parameters play a crucial role in achieving high quality light field reconstruction. As demonstrated in Fig.~\ref{fig:anchor_basic}, truncating parameters outside the interval [-1, 1] leads to severe image degradation, underscoring the significance of preserving these parameters with high precision. Given their relatively small number, we implement 16-bit uniform quantization (half precision in PyTorch) for these parameters, resulting in a negligible increase in the final average bitrate.

While for parameters within the interval [-1, 1], we employ a non-uniform quantization approach using an optimized quantizer. Specifically, given a target centroid number $N$, we apply the \textit{k-means} algorithm to cluster these parameters into $N$ centroids, resulting in a global codebook denoted as $C$. It's worth noting that, unlike the approach taken in \cite{fan2020training}, where authors develop individual local codebooks for each layer of the network, we utilize a single codebook for all layers, and the differences of their performance are discussed in Sec.~\ref{sec:quant_stratgies}. 
Once we have the codebook $C$, we quantize the parameters within the interval [-1, 1] by mapping their values to the nearest centroids. However, due to the interdependencies between layers in the network (the input of the current layer is the output of the precedent layer), quantizing parameters simultaneously across all layers could lead to quantization error accumulation, primarily affecting the quality of the output. To address this issue, we implement a `quantization-aware training' strategy, wherein network layers are quantized one by one. Specifically, after quantizing ${\bm{Q}^{1}_{i'}, \bm{Q}^{2}_{i'}}$ of a certain layer $\mathbb{L}{i'}$, we freeze the weights of all previous layers, denoted as $\mathbb{L}{i}$ for $i \le i'$, and continue training the subsequent layers $\mathbb{L}_{i} \forall i>i'$ as follows:
\begin{gather}
    argmin_{\{\bm{W}_{i},\bm{b}_{i}\}} ||c'-c_{gt}||^{2}_{2}, \forall i>i',\\
    s.t. \ \ \{\bm{\hat{Q}}^{1}_{i}, \bm{\hat{Q}}^{2}_{i}\} \textit{ are fixed}, \forall i \le i'
\end{gather}
where $\{\bm{\hat{Q}}^{1}_{i}, \bm{\hat{Q}}^{2}_{i}\}$ are non-uniformly quantized parameters. 
In this manner, our quantization approach allows for the compensation of quantization errors through the updates of network weights. Fig.~\ref{fig:weight_distribution_after} displays the weight distribution after rate-constrained quantization with 256 centroids. In Fig.~\ref{fig:quantization}, we present the observed performance evoluton as we sequentially quantize layers from the first to the last. It's important to note that our quantization-aware training procedure effectively minimizes PSNR loss in the initial layers. The primary source of performance degradation arises from the quantization of specific layers and remains relatively limited.

\subsubsection{Huffman coding}
In addition to the quantization operation performed with an optimized quantizer, further reductions in bitrate can be achieved by employing Huffman coding \cite{huffman1952method}. However, a challenge arises due to the differing quantization schemes applied to parameters within the interval [-1, 1] and those outside this range. Specifically, parameters within the interval are non-uniformly quantized, while those outside are quantized using 16-bit fixed-point scalars. If we were to apply Huffman coding to these two sets of parameters separately, an additional flag bit would be required to signal to which set a transmitted parameter belongs. The combined entropy of these two sets can be roughly estimated as $\mathcal{H}(S_{1})+\mathcal{H}(S_{2})+1$, where $\mathcal{H}$ represents the entropy of the parameter set, and $S_{1}$ and $S_{2}$ represent the sets of quantized parameters inside and outside [-1,1], respectively. It is evident that the inclusion of the extra flag bit would significantly impact the final bitrate.

To save on the expense of using a flag bit, we've adopted a streamlined approach. Instead of creating separate Huffman code tables for parameters within and outside the [-1, 1] interval, we've created a single code table that covers both sets. The length of this combined table is the sum of the centroid count $N$, and the number of parameters outside [-1, 1]. While encoding both sets together theoretically increases entropy as $\mathcal{H}(\{S_1, S_2\}) > \mathcal{H}(S_1) + \mathcal{H}(S_2)$, it's more efficient in terms of bitrate as it eliminates the need for the flag bit, leading to $\mathcal{H}(\{S_1, S_2\}) < \mathcal{H}(S_1) + \mathcal{H}(S_2) + 1$.

\subsection{Network Distillation}
\label{sec:distillation}

\subsubsection{Conflict between low rank approximation and quantization}

As previously mentioned, both low-rank approximation and rate-constrained quantization serve the purpose of compacting the NeRF representation, but they target different aspects—parameter count and per-parameter bit count. It might seem intuitive to incorporate both low-rank approximation and quantization within the ADMM framework, constraining both rank and bit count. However, due to the non-convex nature of these functions, attempting to optimize them simultaneously within the ADMM framework can lead to convergence challenges.

An alternative approach involves performing low-rank approximation and quantization consecutively. First, we fine-tune the network with a low-rank constraint to obtain $\bm{W}^{r}_{i}$. Next, we decompose $\bm{W}^{r}_{i}$ into TT components, denoted as $\{\bm{Q}^{1}_{i},\bm{Q}^{2}_{i}\}$, and subsequently quantize them to obtain $\{\hat{\bm{Q}}^{1}_{i},\hat{\bm{Q}}^{2}_{i}\}$. This sequence of operations can be summarized as: $\bm{W}_{i}\to\bm{W}^{r}_{i}\to\{\bm{Q}^{1}_{i},\bm{Q}^{2}_{i}\}\to\{\hat{\bm{Q}}^{1}_{i},\hat{\bm{Q}}^{2}_{i}\}$.
However, this approach faces a challenge. During the rate-constrained quantization step, we freeze the weights of certain layers and update the weights $\bm{W}^{r}_{i}$ in all consecutive layers to compensate for quantization errors. Unfortunately, these updated weights may no longer adhere to the low-rank property. Therefore, one of the challenges in our pipeline design is to achieve quantization while preserving the low-rank characteristics of the network weights.

\begin{figure*}[ht]
    \centering
    \setlength{
    \tabcolsep}{1pt}
    \centering
    \begin{tabular}{ccccc}
\includegraphics[width=0.2\linewidth]{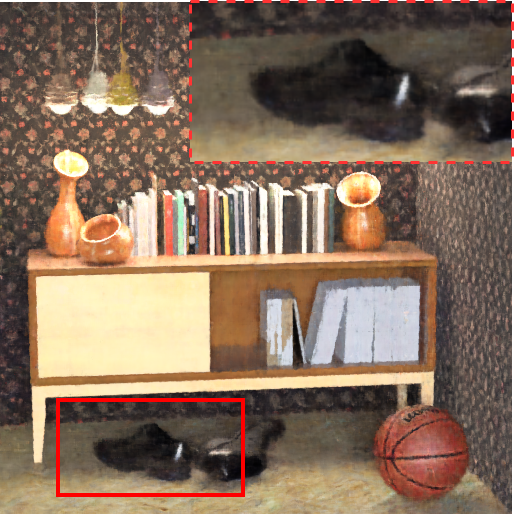} &
\includegraphics[width=0.2\linewidth]{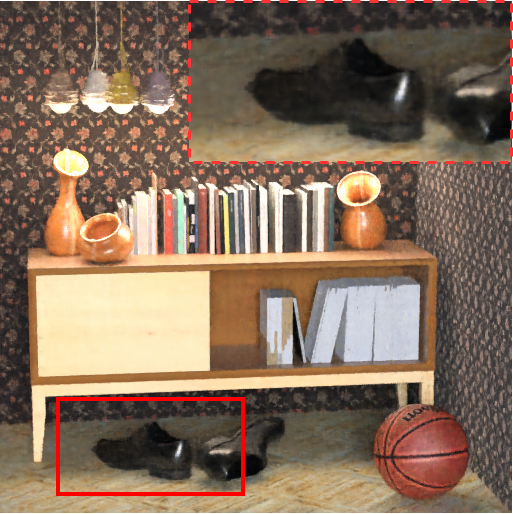} &
\includegraphics[width=0.2\linewidth]{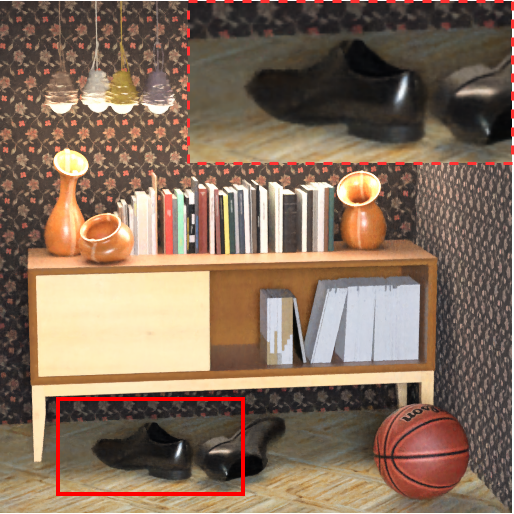} &
\includegraphics[width=0.2\linewidth]{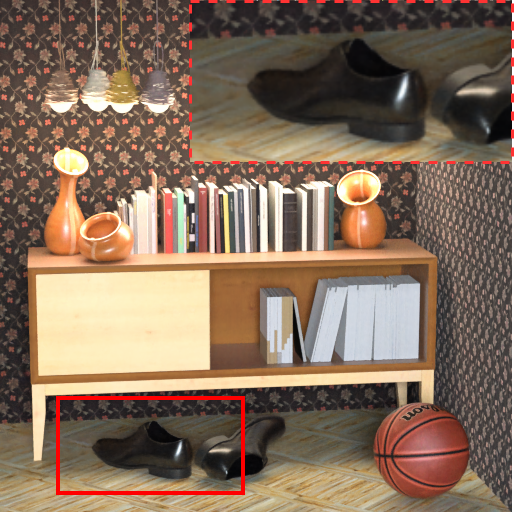} &
\includegraphics[width=0.2\linewidth]{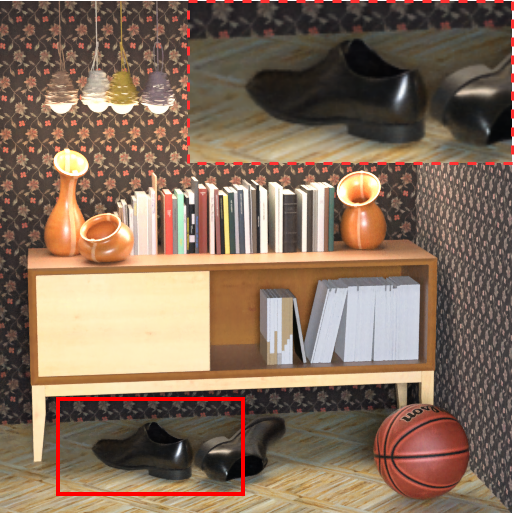} \\

\includegraphics[width=0.2\linewidth]{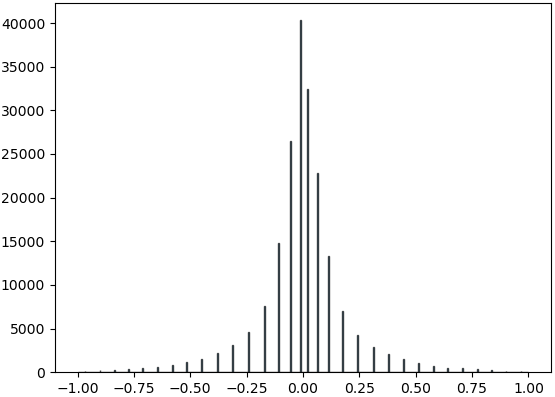} &
\includegraphics[width=0.2\linewidth]{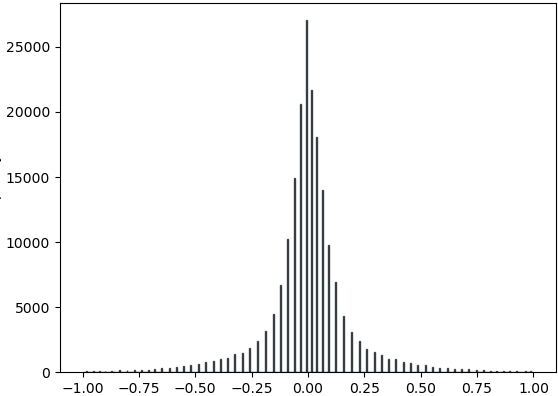} &
\includegraphics[width=0.2\linewidth]{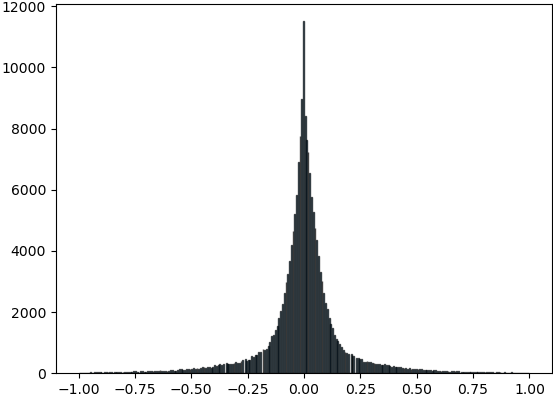} &
\includegraphics[width=0.2\linewidth]{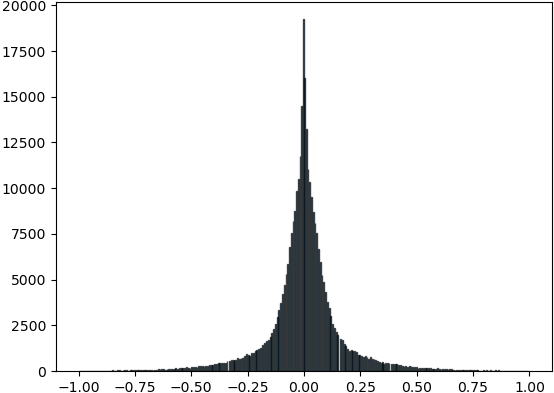} &
\includegraphics[width=0.2\linewidth]{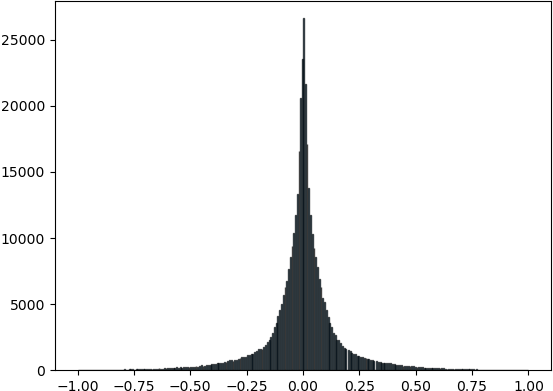}\\
$(r=40,N=32)$ & $(r=40,N=64)$&$(r=40,N=256)$ &$(r=70,N=256)$ & $(r=90,N=256)$\\
PSNR = 24.60dB & PSNR = 27.17dB & PSNR = 31.10dB & PSNR = 34.71dB & PSNR = 35.22dB \\
\end{tabular}
\caption{We show in the first row the central view of the light field `\textit{sideboard}' reconstructed by quantized DLR-NeRF with different rank\&centroid settings. The second row shows the weight distribution of each quantized DLR-NeRF. Larger is the rank of the model and higher is the number of centroids, better will be the reconstructed view.} 
\label{fig:rank_centroid}
\end{figure*}

\subsubsection{Construction of DLR-NeRF}
To tackle this challenge, we insert a network distillation step between the low-rank approximation and quantization stages. This step involves creating a slimmer network, DLR-NeRF, from the original LR-NeRF network.

More specifically, after the low-rank approximation step, we decompose the weights $\bm{W}^{r}_{i}$ in each layer $\mathbb{L}_{i}$ of LR-NeRF into TT components $\{\bm{Q}^{1}_{i},\bm{Q}^{2}_{i}\}$. If $\{\bm{Q}^{1}_{i},\bm{Q}^{2}_{i}\}$ has fewer parameters than $\bm{W}^{r}_{i}$, i.e., $(n_{1}+n_{2})\times r < (n_{1}\times n_{2})$, we initialize two cascaded fully-connected layers $\mathbb{L}^{'1}_{i}$ and $\mathbb{L}^{'2}_{i}$ in DLR-NeRF with $\bm{Q}^{1}_{i}$ and $\{\bm{Q}^{2}_{i},\bm{b}_{i}\}$. Otherwise, we initialize one layer $\mathbb{L}^{'}_{i}$ in DLR-NeRF with $\{\bm{W}^{r}_{i},\bm{b}_{i}\}$. Consequently, DLR-NeRF can be seen as a deeper version of LR-NeRF with fewer layer parameters.

We borrow the term `distillation' from network compression field to describe the process, as initializing DLR-NeRF with TT components from LR-NeRF is akin to creating a student network (DLR-NeRF) from a teacher network (LR-NeRF). The student network produces similar outputs but has fewer parameters.
Importantly, the layer $\mathbb{L'}$ of DLR-NeRF not only stores scene information from LR-NeRF but also maintains its low-rank shape ($n_{1}\times r$ or $r\times n_{2}$), unaffected by subsequent quantization. The fine-tuning of DLR-NeRF employs the same loss function as Eq.~\ref{eq:loss} to update its parameters.

The network distillation serves as a bridge between low-rank approximation and quantization. The distilled DLR-NeRF can undergo further quantization, as detailed in Sec.~\ref{sec:quantization}, without compromising the low-rank property of its weights.
It's worth noting that one could choose to skip the NeRF initialization and low-rank finetuning steps and train a distilled NeRF from scratch. However, as discussed in Sec.~\ref{sec:random_init}, weights trained from scratch follow a different distribution from low-rank weights and may not produce high-quality scene reconstructions.

After applying the quantization operation on this DLR-NeRF, the obtained QDLR-NeRF can be used to represent a light field in the compression context. One can indeed transmit the quantized network weights $\hat{\bm{Q}}^{1}_{i}, \hat{\bm{Q}}^{2}_{i}$ (after Huffman Coding), $\hat{\bm{b}}_{i}$ (16 bits), the codewords (32 bits) and the camera parameters $f,\Delta$ (32 bits) from the sender to the receiver.

\begin{figure*}[!ht]
    \centering
    \setlength{
    \tabcolsep}{1pt}
    \centering
    \begin{tabular}{cc}
\includegraphics[width=0.43\linewidth,height=0.23\textheight]{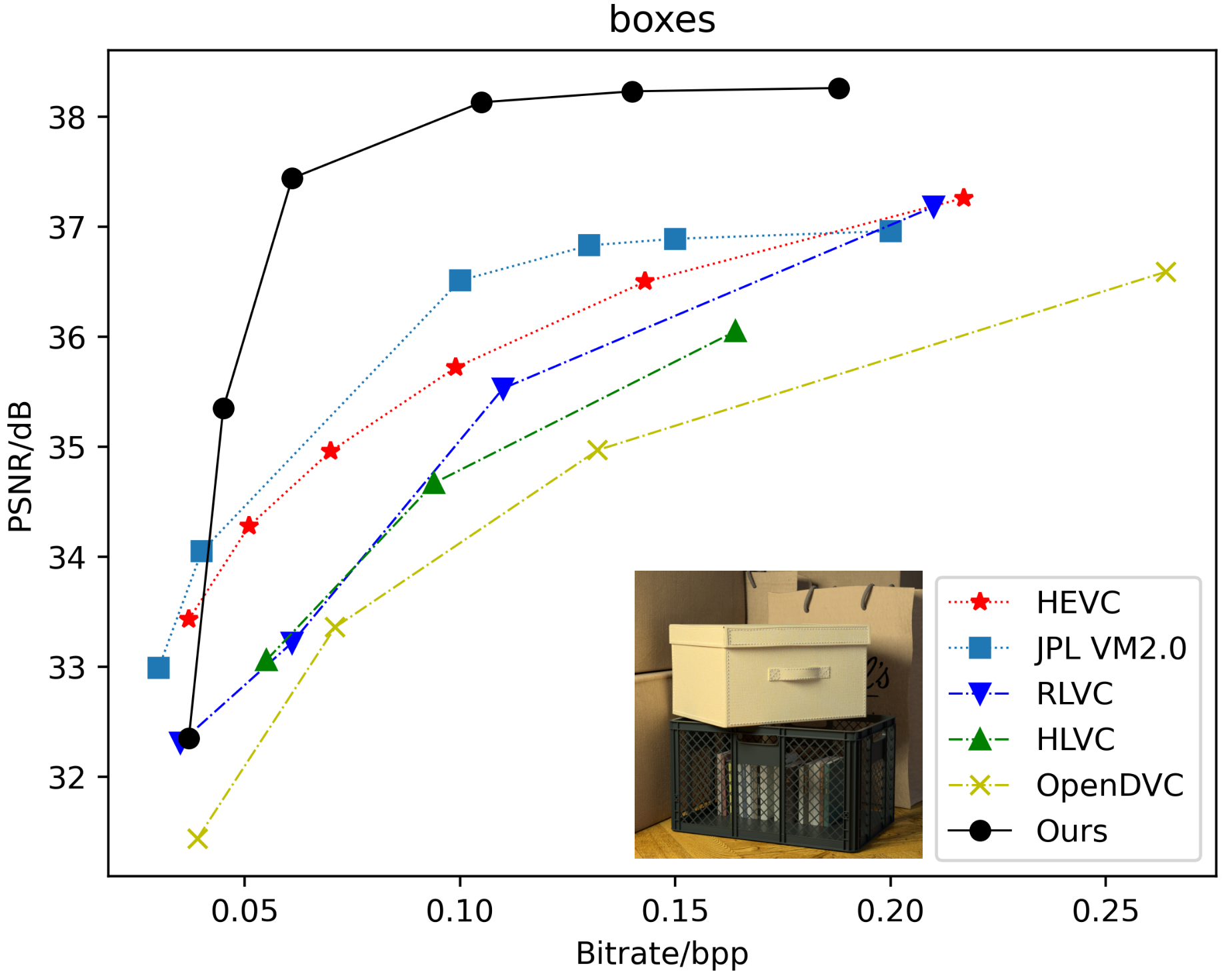} &
\includegraphics[width=0.43\linewidth,height=0.23\textheight]{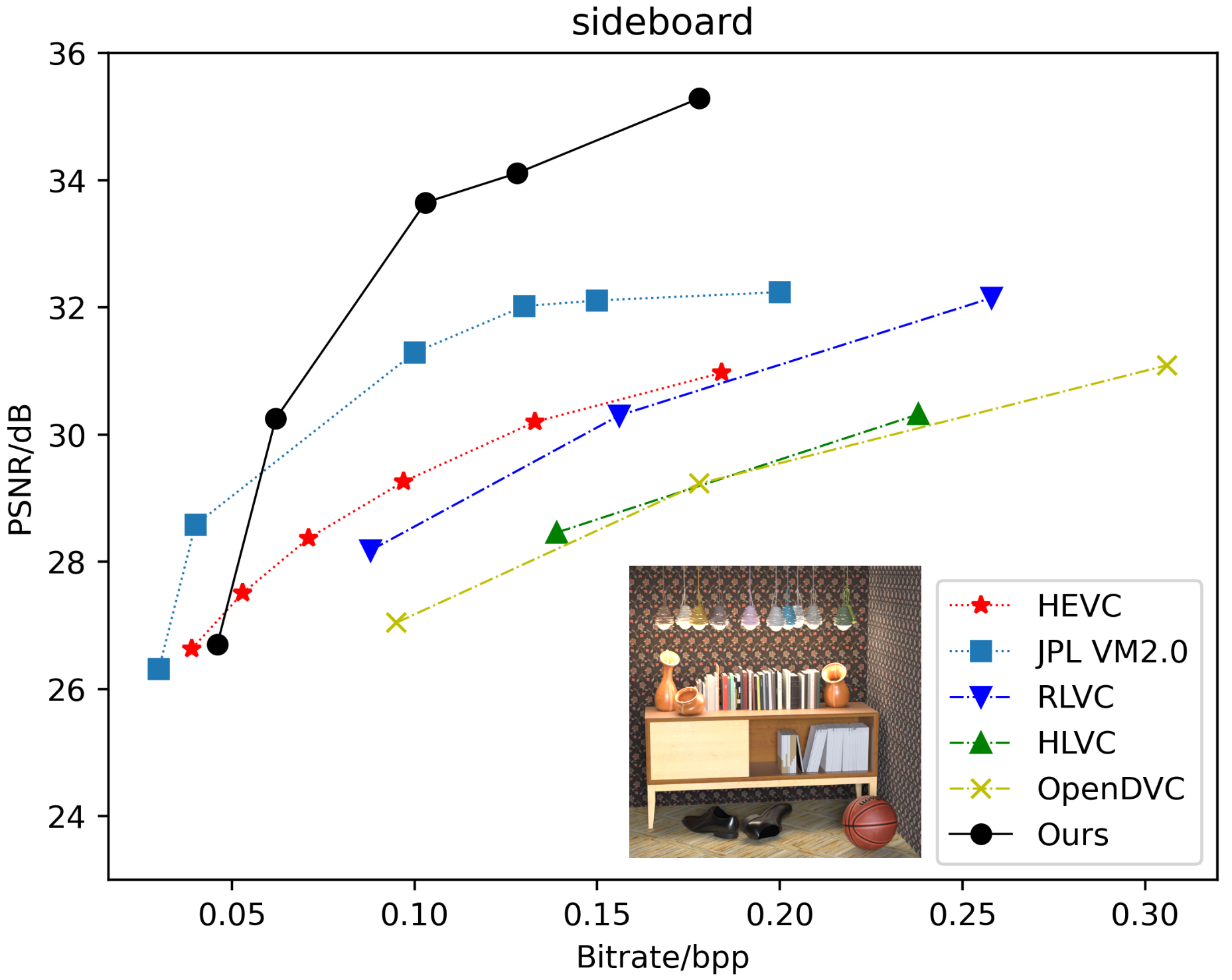} \\
\includegraphics[width=0.43\linewidth,height=0.23\textheight]{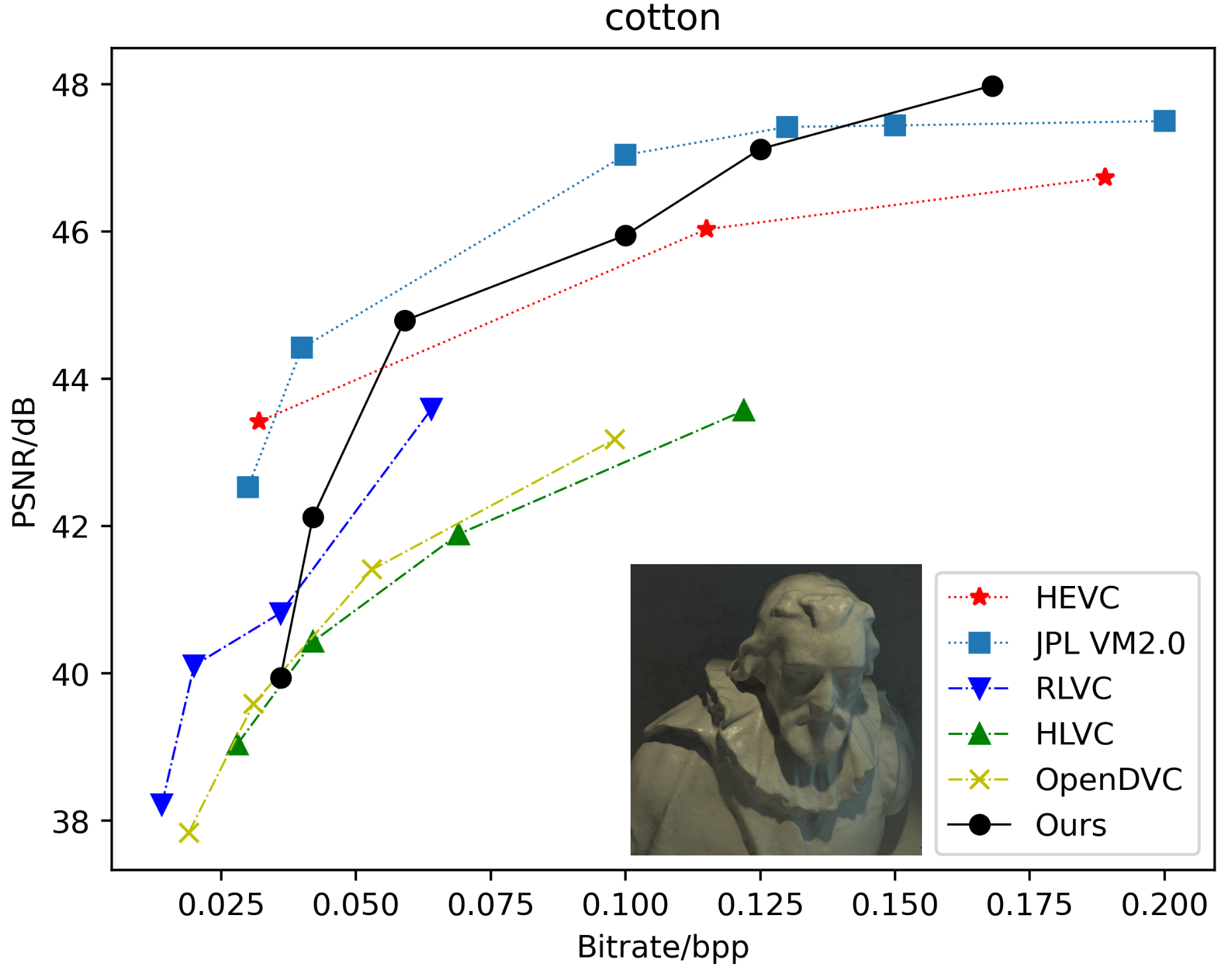} &
\includegraphics[width=0.43\linewidth,height=0.23\textheight]{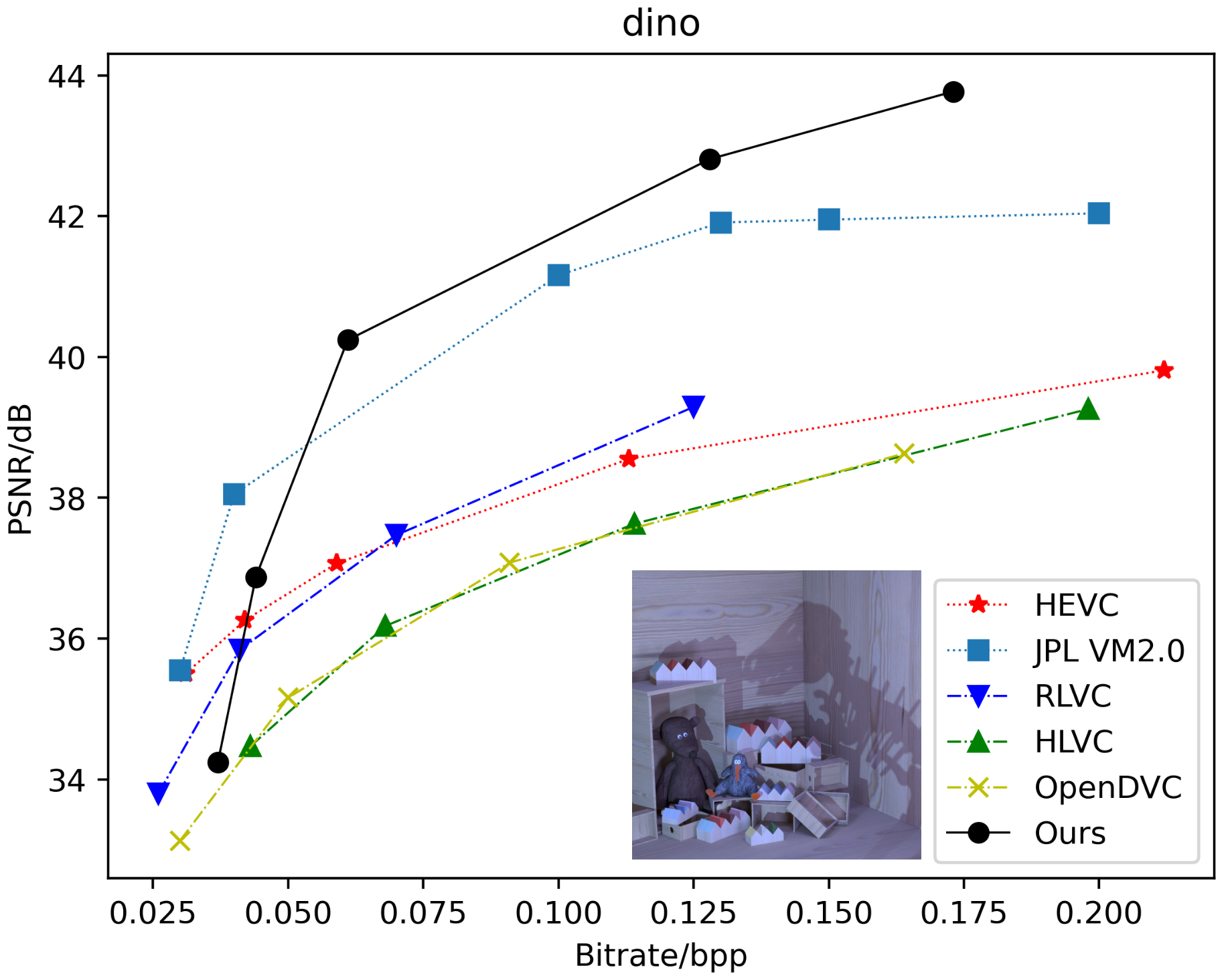} \\
\includegraphics[width=0.43\linewidth,height=0.23\textheight]{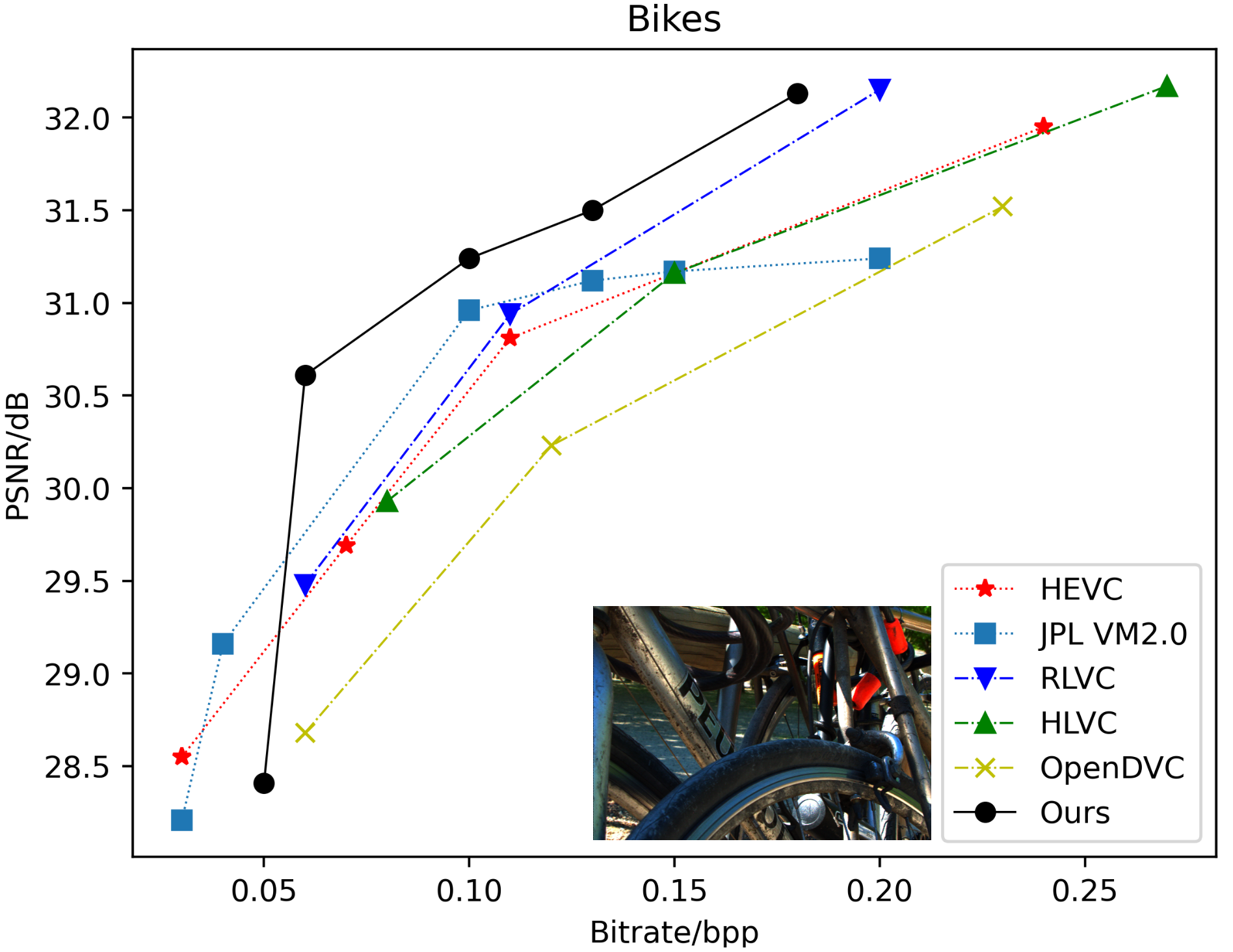} &
\includegraphics[width=0.43\linewidth,height=0.23\textheight]{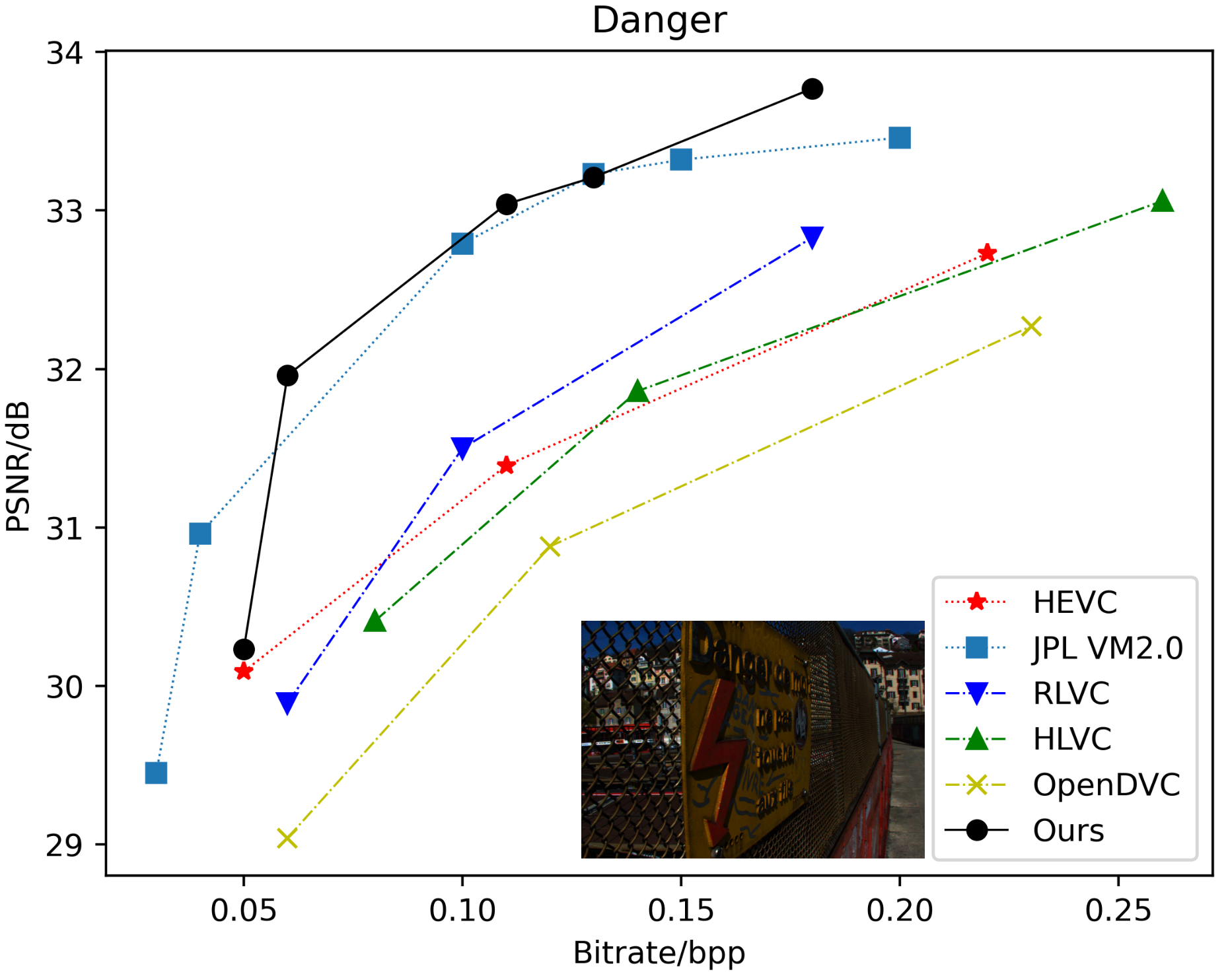} \\
\includegraphics[width=0.43\linewidth,height=0.23\textheight]{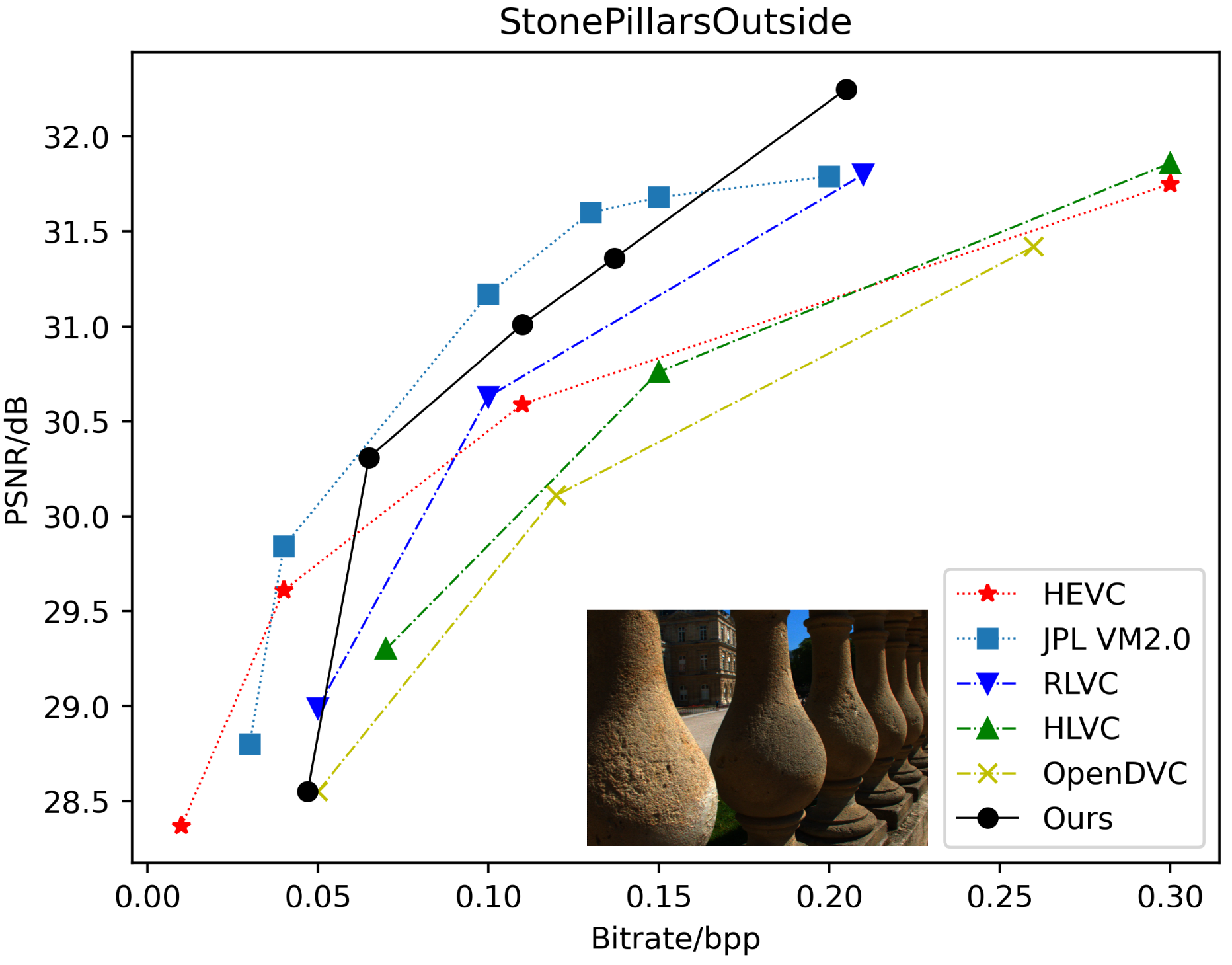} &
\includegraphics[width=0.43\linewidth,height=0.23\textheight]{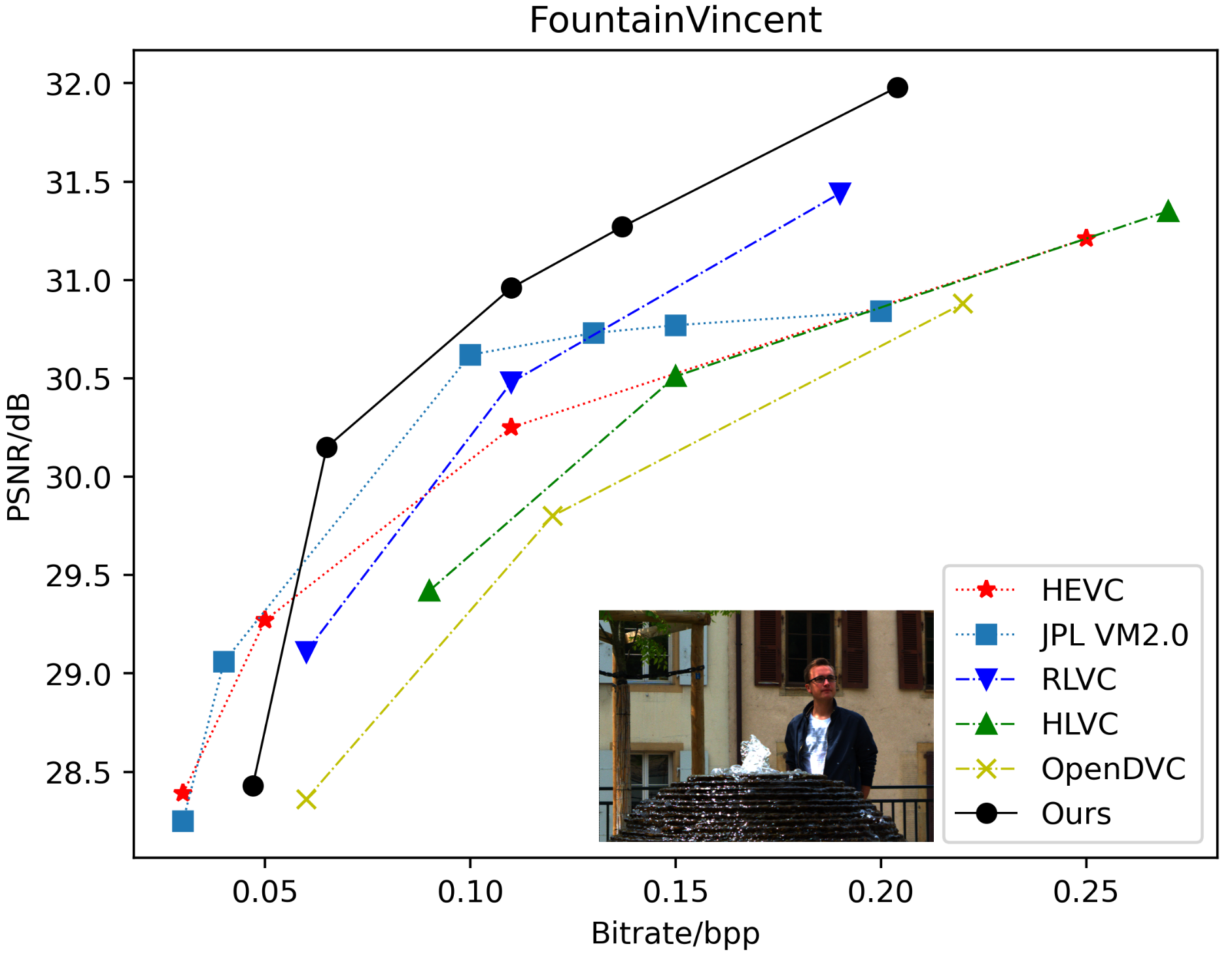} \\
\end{tabular}
\caption{Rate-distortion curves of different compression methods, with tested light fields from the HCI synthetic dataset \cite{honauer2016dataset} and the EPFL real-world dataset \cite{EPFLLFdataset}.} 
\label{fig:rate_distortion}
\end{figure*}

\section{Training schedule and hyperparameters}
The training of such a workflow has four phases,
as follows.

a).\textbf{Training the simplified NeRF:} We simplified the architecture of NeRF by adopting one single MLP for inference. The depth of the adopted MLP is $8$ and the number of channels is 256. We set the same query point number $128$ along each camera ray as in \cite{mildenhall2020nerf},
and the camera pose parameters $(f,\Delta)$ are initialized to $f= 0.01W$ ($W$ is the width of the light field views), $\Delta=10$. We adopt normalized depth for rendering. When training such a network, we followed the default configuration in \cite{mildenhall2020nerf}, with the initial learning rate set to $5\times 10^{-4}$ and the exponential decay rate to $0.1$. We finally trained the network for $3\times 10^{5}$ iterations.

b).\textbf{Finetuning LR-NeRF:} We finetuned LR-NeRF following the same configuration as in the previous step, except that in this step, the camera pose parameters are no longer updated. We finetune LF-NeRF for $3\times 10^{5}$ iterations. The penalty factor $\rho$ for ADMM is set to $10$.

c).\textbf{Finetuning DLR-NeRF:} After the TT decomposition of the LR-NeRF weights and the initialization of DLR-NeRF, we continued to finetune DLR-NeRF for $2\times 10^{5}$ iterations, during which the camera pose parameters are kept fixed.

d).\textbf{Quantizing DLR-NeRF:} The quantization of the layers of DLR-NeRF is gradually carried out from the first to the last layer. 
We train the other layers for $3 \times 10^{4}$ iterations after quantizing each layer. The initial codebook for the interval $[-1,1]$ is obtained after $20$ iterations of clustering in the k-means algorithm. Our proposed workflow has been implemented using pytorch and has been trained on a GPU of type GeForce RTX 2080 Ti with 11 GB memory. 

\begin{figure*}[!ht]
    \centering
    \setlength{
    \tabcolsep}{1pt}
    \centering
    \begin{tabular}{cccccc}
GT view(5,5) & RLVC & HLVC & HEVC & JPL & Ours\\
\includegraphics[width=0.16\linewidth]{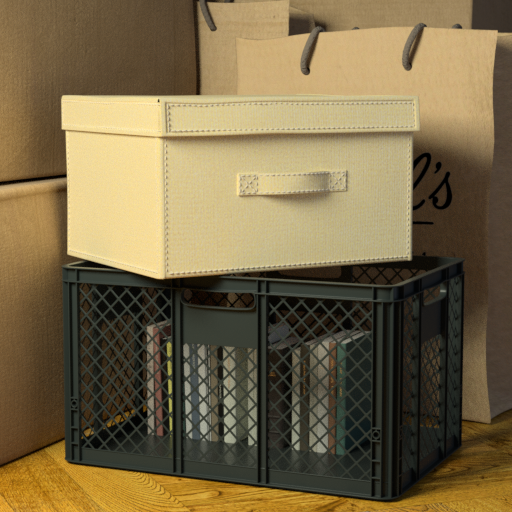} &
\includegraphics[width=0.16\linewidth]{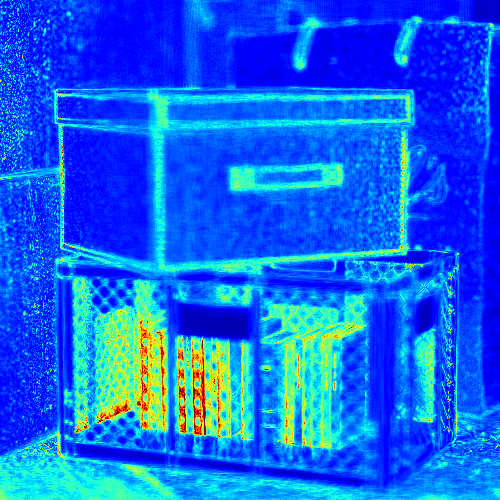} &
\includegraphics[width=0.16\linewidth]{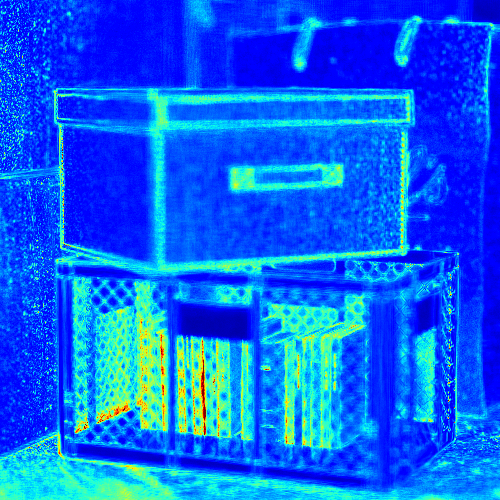} &
\includegraphics[width=0.16\linewidth]{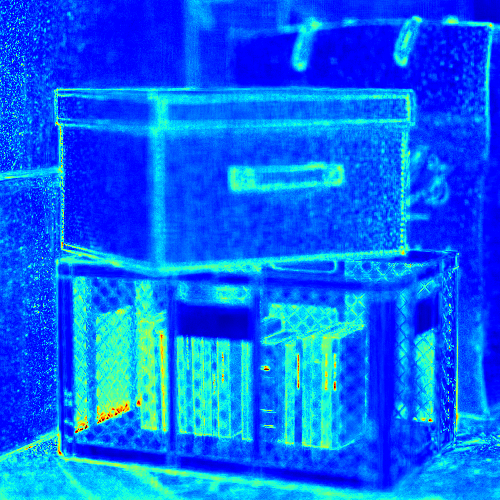} &
\includegraphics[width=0.16\linewidth]{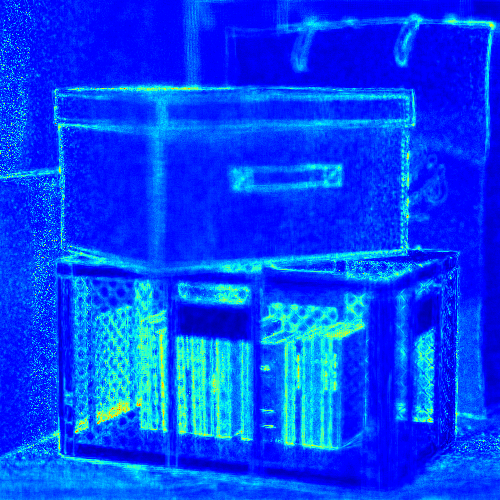} &
\includegraphics[width=0.16\linewidth]{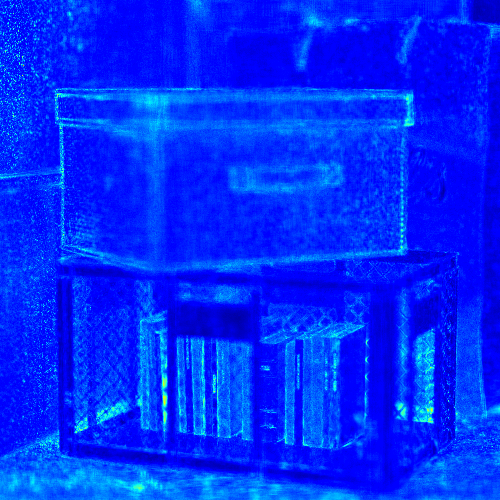}\\
boxes & PSNR=35.49dB & PSNR=34.68dB & PSNR=35.73dB & PSNR=36.51 & PSNR=38.19dB\\
& bpp=0.110 & bpp=0.094 & bpp=0.099 & bpp=0.100 & bpp=0.105\\
\includegraphics[width=0.16\linewidth]{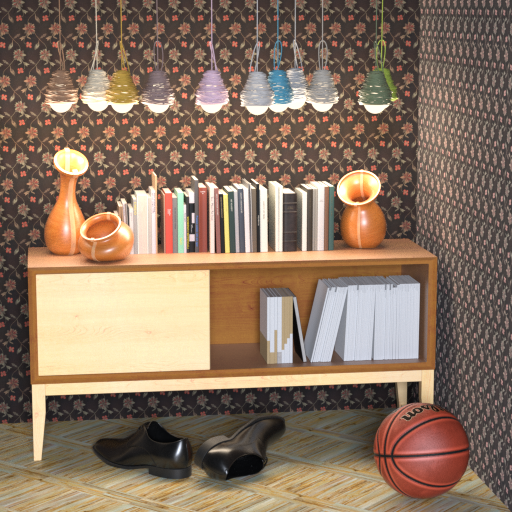} &
\includegraphics[width=0.16\linewidth]{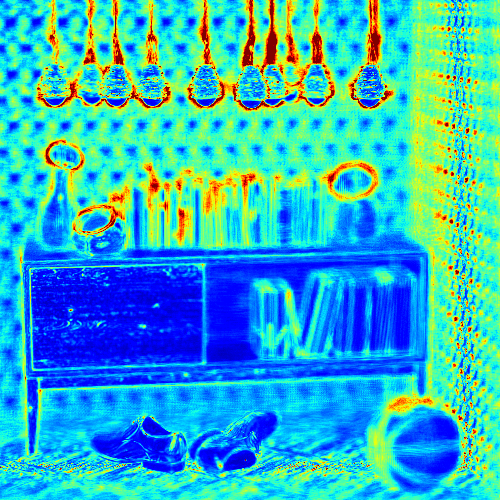} &
\includegraphics[width=0.16\linewidth]{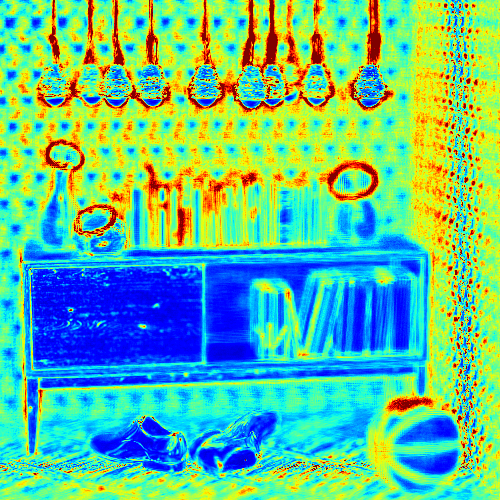} &
\includegraphics[width=0.16\linewidth]{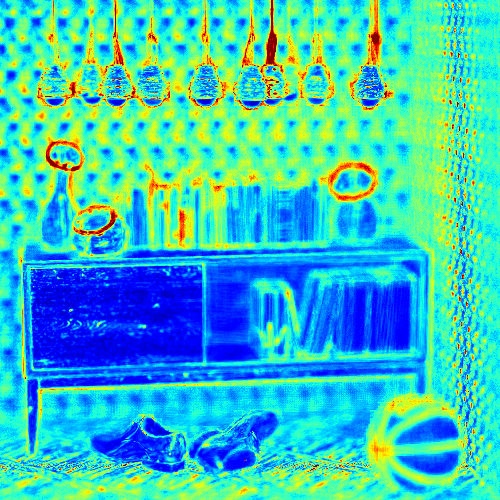} &
\includegraphics[width=0.16\linewidth]{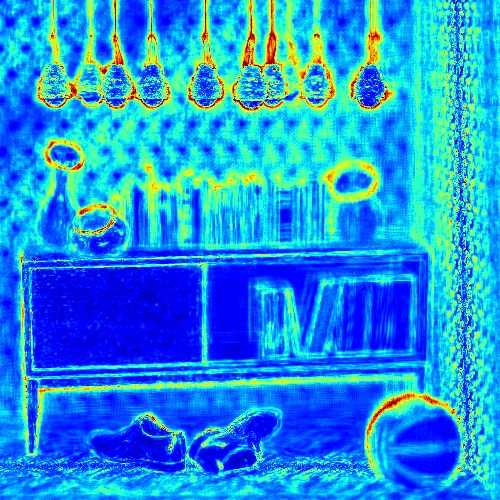} &
\includegraphics[width=0.16\linewidth]{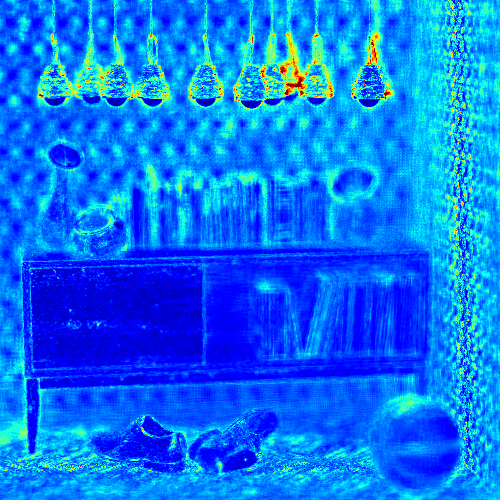}\\
sideboard & PSNR=30.36dB & PSNR=28.52dB & PSNR=30.21dB & PSNR=32.02 & PSNR=34.18dB\\
& bpp=0.156 & bpp=0.139 & bpp=0.133 & bpp=0.130 & bpp=0.128\\
\includegraphics[width=0.16\linewidth]{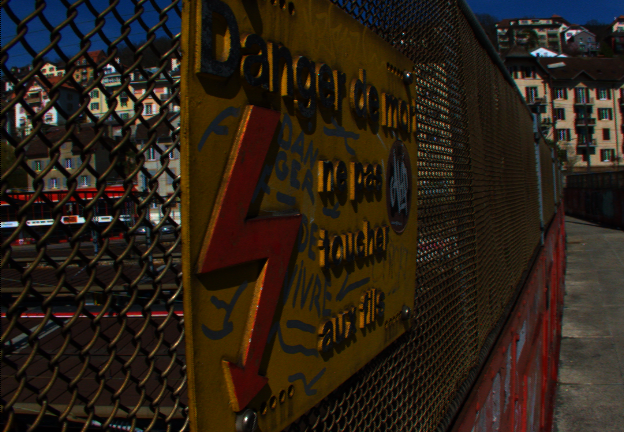} &
\includegraphics[width=0.16\linewidth]{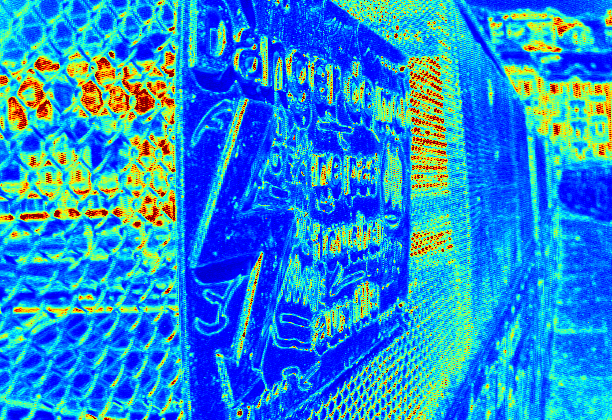} &
\includegraphics[width=0.16\linewidth]{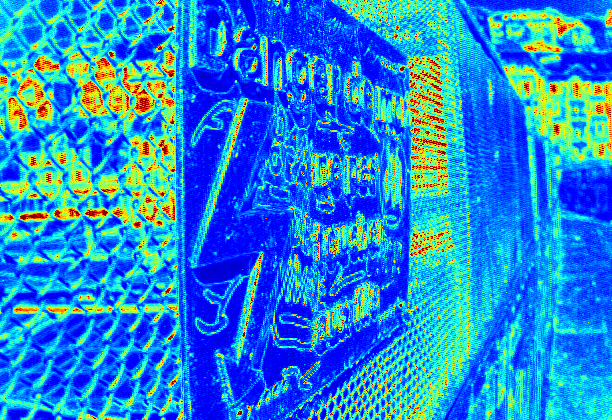} &
\includegraphics[width=0.16\linewidth]{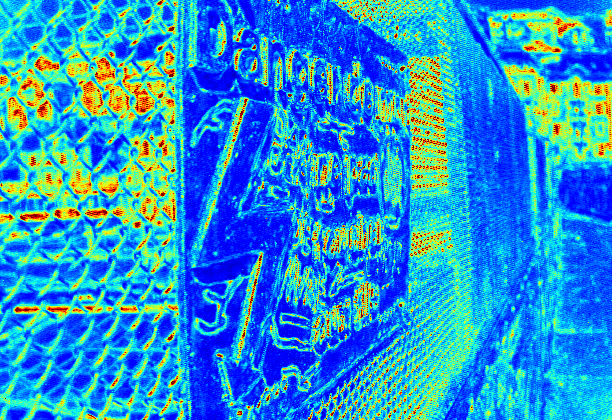} &
\includegraphics[width=0.16\linewidth]{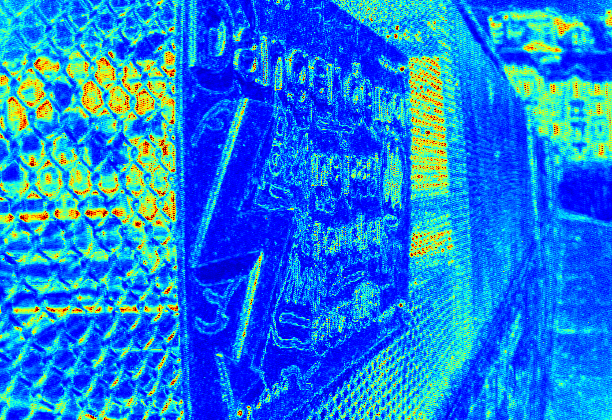} &
\includegraphics[width=0.16\linewidth]{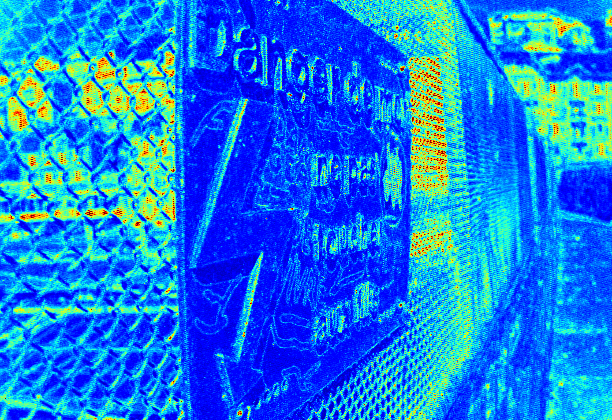}\\
Danger & PSNR=31.50dB & PSNR= 31.86dB & PSNR=31.39dB & PSNR=32.79 & PSNR=33.04dB\\
& bpp=0.104 & bpp=0.141 & bpp=0.110 & bpp=0.100 & bpp=0.110\\
\includegraphics[width=0.16\linewidth]{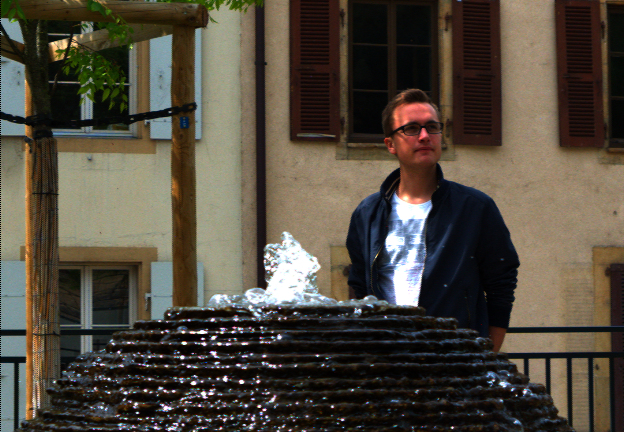} &
\includegraphics[width=0.16\linewidth]{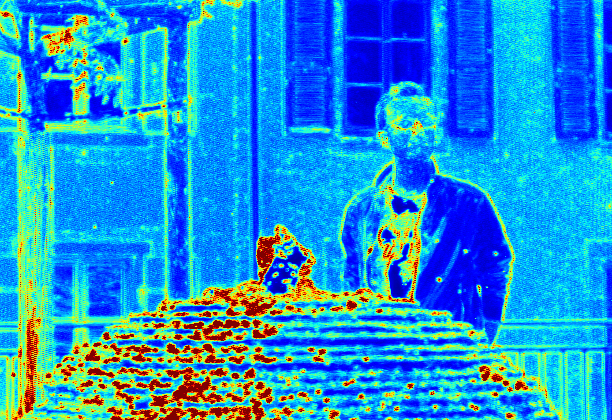} &
\includegraphics[width=0.16\linewidth]{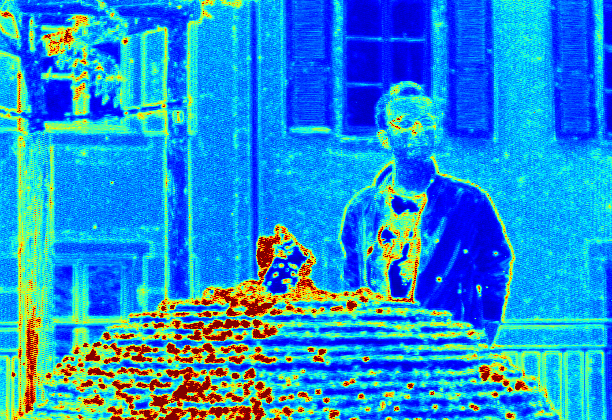} &
\includegraphics[width=0.16\linewidth]{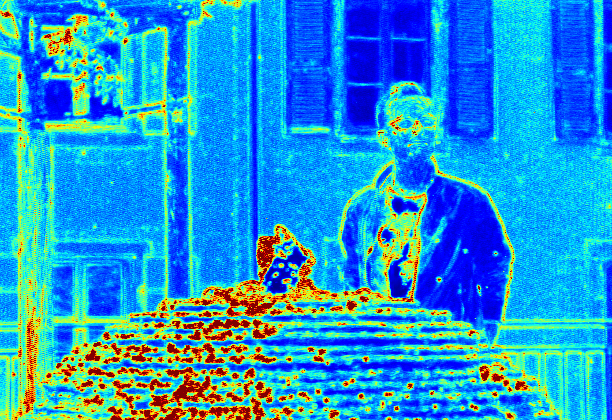} &
\includegraphics[width=0.16\linewidth]{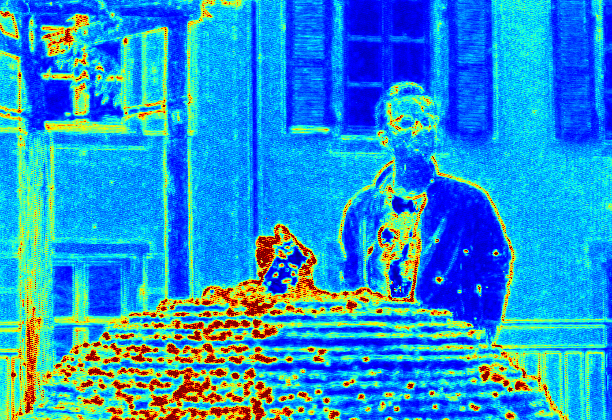} &
\includegraphics[width=0.16\linewidth]{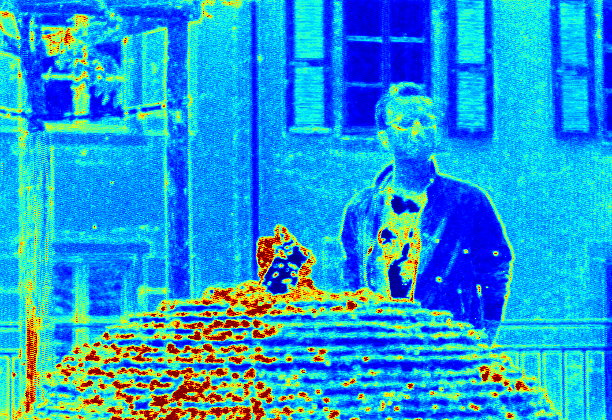}\\
FountainVincent & PSNR=30.48dB & PSNR= 30.51dB & PSNR=30.25dB & PSNR=30.62 & PSNR=30.96dB\\
& bpp=0.110 & bpp=0.090 & bpp=0.110 & bpp=0.100 & bpp=0.103\\
\end{tabular}
\caption{Averaged reconstruction error maps of decompressed light fields, when using the RLVC, HLVC, HEVC, Jpeg-pleno methods and our method, along with the corresponding PSNR and bitrate values.}
\label{fig:error_map}
\end{figure*}

\section{Compression performance analysis}
\subsection{Settings}
We first trained a simplified NeRF with full rank. This initial model is then finetuned and distilled using four different ranks $r = \{150, 90, 70, 40\}$, and we finally quantized the distilled models with $N =256$ centroids.
The quantized models having a lower rank have fewer parameters, hence give a lower bitrate. We found that, when the target rank $r$ is lower than $40$, the light field reconstructed with the distilled Low-Rank NeRF (DLR-NeRF) is prone to artifacts.
Therefore, we do not decrease the rank below $40$, and to decrease the bitrate, we reduce the number of centroids to further reduce the model size.
More precisely, we set $N = \{256, 64, 32\}$ when $r = 40$. 

To test the effectiveness of our method, we carry out experiments using both synthetic and real-world light fields. Both of them are from widely used light field benchmark. For synthetic light fields, we took four scenes from the HCI dataset \cite{honauer2016dataset}, i.e. \textit{boxes, sideboard, cotton, dino}, each of them has an angular resolution $(U,V)=(9,9)$ and a spatial resolution $(X,Y)=(512,512)$. While for real-world light fields, we took four scenes \textit{Bikes, Danger, StonePillarsOutside, FountainVincent2} from the EPFL light field dataset \cite{EPFLLFdataset}, which have been captured using a plenoptic Lytro Illum camera. We took the central $9\times9$ views, each view having a spatial resolution of $432\times 624$. Although ground truth camera poses are already available for synthetic data, we estimated the camera poses for both the synthetic and real-world data to show that our workflow is not limited by the availability of the camera pose parameters.

\subsection{Rate-Distortion Performance}
Fig.~\ref{fig:rank_centroid} shows the central view reconstructed using the QDLR-NeRF with different $(r,N)$ settings, and their corresponding weight probability distributions. One can easily notice that, larger is the rank of the DLR-NeRF model or higher is the number of centroids better is the reconstructed view. This is quite intuitive, as a larger rank means more parameters a model has, and more centroid means less quantization noise. 

However, in the compression context, besides the quality of the `compressed views', we also need to take bitrate into account, i.e. the rate-distortion performance. To investigate the compression performance of our proposed method, we measure the rate-distortion performances with different rank\&centroid settings, and compare them with those obtained when applying State-Of-The-Art (SOTA) video compression methods to light fields, by encoding the views as a pseudo video sequence. The codecs that we consider are the HEVC-Lozenge \cite{rizkallah2016impact}, the
Jpeg-pleno VM2.0 standard designed specifically for light field compression,
and learning-based RLVC \cite{yang2021learning}, HLVC \cite{yang2020Learning} and OpenDVC \cite{yang2020opendvc} compression methods. OpenDVC is an open implementation of the DVC \cite{Lu2019DVC}.

For the settings of each reference method, we use the HEVC HM-16.10 implementation in our test, with a GOP size equal to $4$. 
The Jpeg-pleno standard needs an input disparity map that we estimated using the technique in \cite{shi2019depth}.
When considering the RLVC, HLVC and OpenDVC methods, we directly used the authors' codes with the recommended configurations, i.e., for the RLVC method, we set the P-frames number to $6$ for both the forward and backward directions, which means a GOP size of $13$, while for the HLVC and OpenDVC methods, we used the default setting of GOP size equal to 10. When testing the reference methods RLVC, HLVC and OpenDVC, we used their trained models. There are four pre-trained models for each method, with an optional hyper-parameter $\lambda = \{256, 512, 1024, 2048\}$ that controls the trade-off between distortion and bitrate. Higher $\lambda$ means less distortion but larger bitrate.

Since our method is based on NeRF, originally designed for view synthesis, we use PSNR on RGB channels to measure reconstruction quality instead of YUV.
Fig.~\ref{fig:rate_distortion} shows the PSNR-Rate curves of each method with different light fields. The bitrate is evaluated by bit per pixel (bpp). One can observe from the figures that our proposed method outperforms the reference methods on synthetic data at moderate bitrate by a margin about 1dB in most of the tested scenes. While on real-world light fields, though the margin is smaller than that of synthetic data, our method still shows superiority than other referenced methods in most of cases. Let us note that, compared to synthetic light fields, real-world light fields captured by Lytro Illum exhibit more noise and various artifacts, including vignetting and blurriness. These factors introduce inconsistencies between views, which can hinder our method's ability to learn accurate scene information and result in lower-quality reconstructed views.

Figure \ref{fig:error_map} shows the averaged error maps for the different methods. Note that the error maps are averaged over all the views, to account for the fact that some methods, such as Jpeg-Pleno, RLVC, OpenDVC, lead to strong variations in quality on the different views (see Fig.~\ref{fig:psnr_indices}).
We can observe that our method yields lower error and better compression results than other methods on synthetic data (\textit{boxes}, \textit{sideboard}). When using real-world Lytro light fields suffering from color inconsistency between angular views, our method may learn inaccurate scene information and hence generate some errors (like the windows in scene \textit{FountainVincent}). However, it can still well capture scene geometry information and better reconstruct the object contours (like the contours of the person in the scene).

Compared with the referenced methods that compress each input frame, our method focuses on retrieving the entire scene with as fewer parameters as possible. This distinct design endows our method with two advantages: 

\begin{figure}
    \centering
  \includegraphics[width=0.99\linewidth]{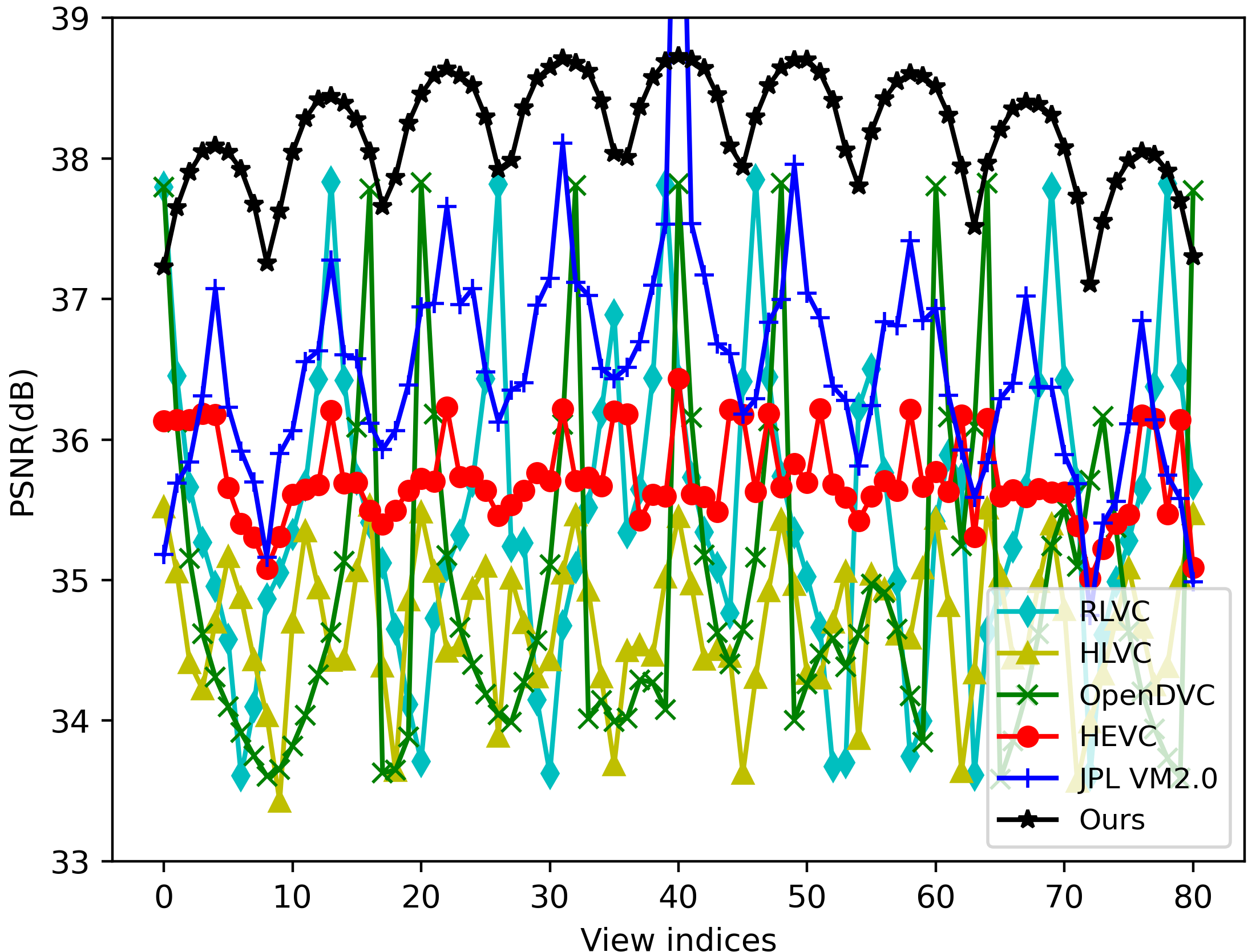}
  \caption{PSNR variation along the viewpoints for the light field `\textit{boxes}'. All views have been compressed at a similar bitrate.}
  \label{fig:psnr_indices}
\end{figure}

1). Better consistency across views. 
Fig.~\ref{fig:psnr_indices} displays PSNR variations based on viewpoints. We can observe that methods like RLVC, HLVC, and OpenDVC initially intra-code key frames and then compress other frames within the GOP using motion-compensated inter-coding. In contrast, Jpeg-pleno employs the central view as the reference for subsequent processes. Views used as key frames or reference tend to have better quality than other decompressed views, resulting in significant quality differences among viewpoints. Unlike methods relying on inter-coding and reference views, our method eliminates the need for a reference view, resulting in more consistent decompression quality across rendered views.

2). Capacity of rendering views in any angular position. Different from other methods that compress a set of input views, our method, thanks to NeRF, encodes the scene information. Hence the scene can be flexibly reconstructed by rendering the light field in any angular position on the decoder side.

\begin{figure*}[!ht]
    \centering
    \setlength{
    \tabcolsep}{1pt}
    \centering
    \begin{tabular}{ccccc}
\includegraphics[width=0.2\linewidth,height=0.2\linewidth]{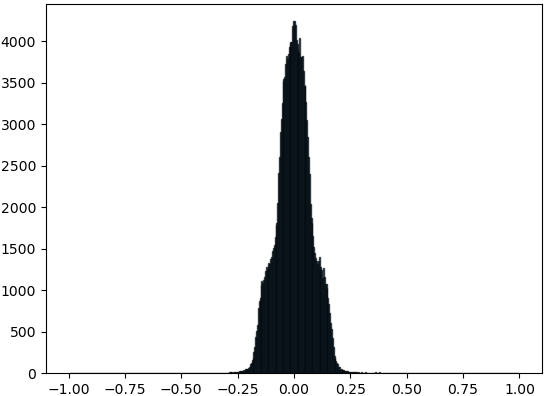} &
\includegraphics[width=0.2\linewidth]{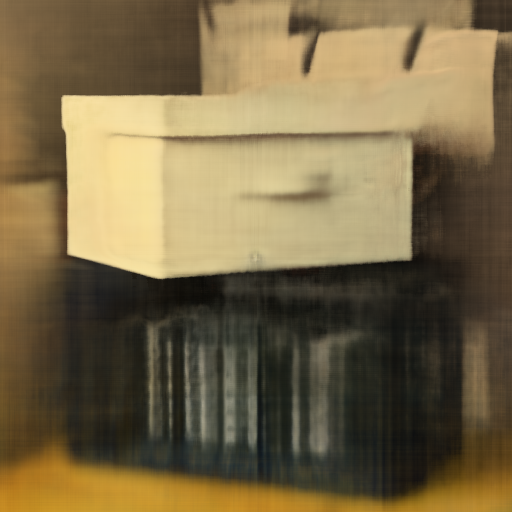} &
\includegraphics[width=0.2\linewidth]{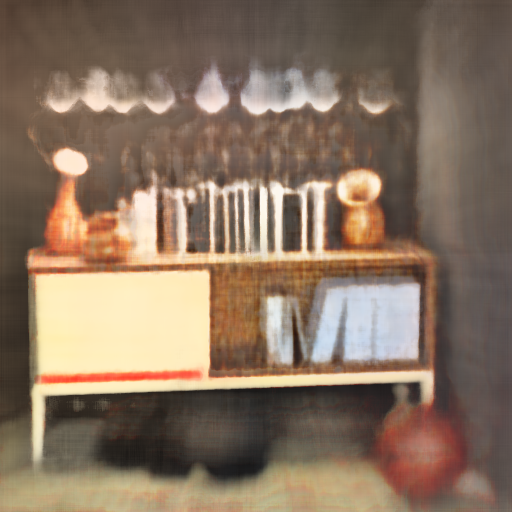} &
\includegraphics[width=0.2\linewidth]{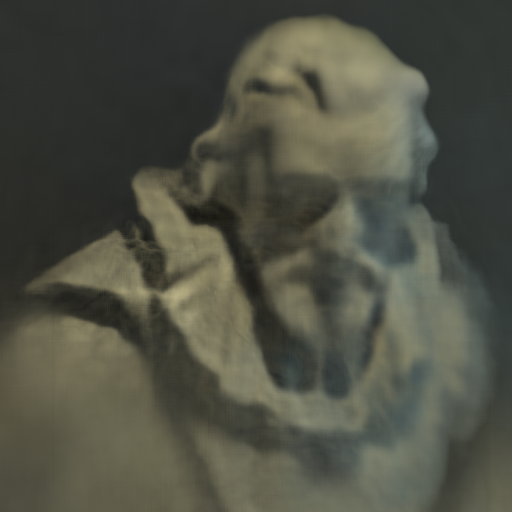} &
\includegraphics[width=0.2\linewidth]{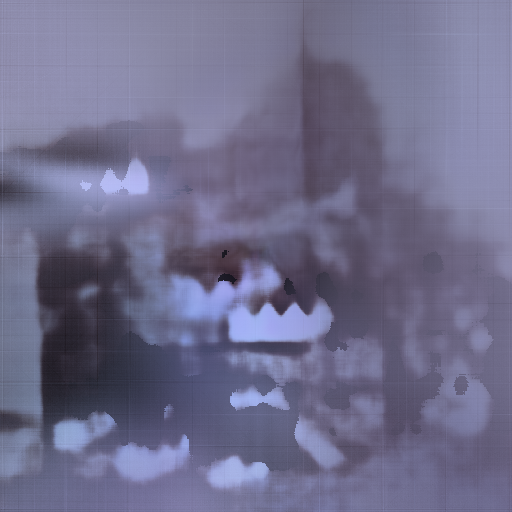} \\
\includegraphics[width=0.2\linewidth,height=0.2\linewidth]{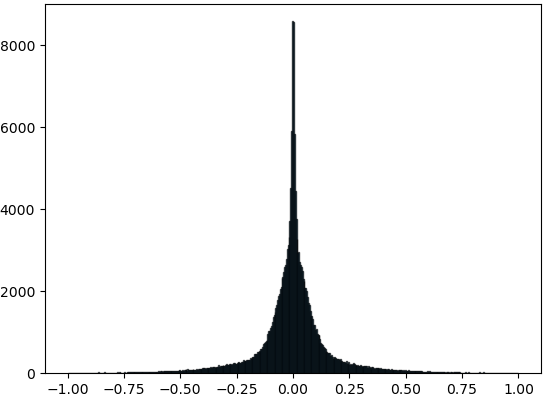} &
\includegraphics[width=0.2\linewidth]{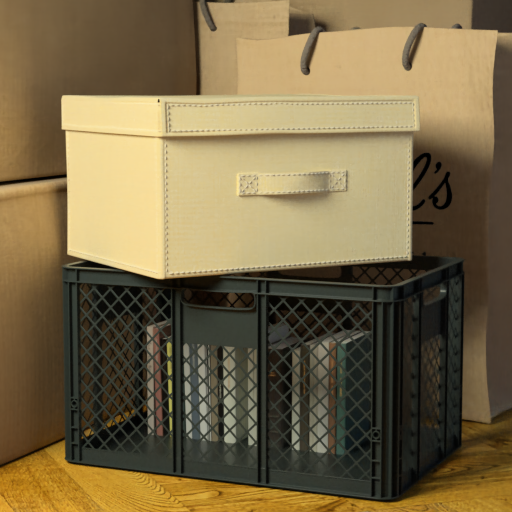} &
\includegraphics[width=0.2\linewidth]{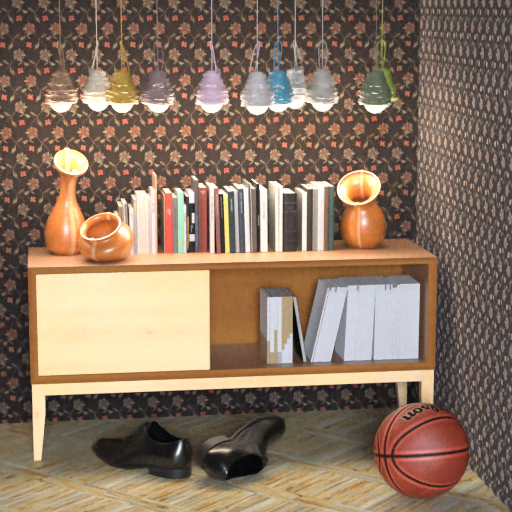} &
\includegraphics[width=0.2\linewidth]{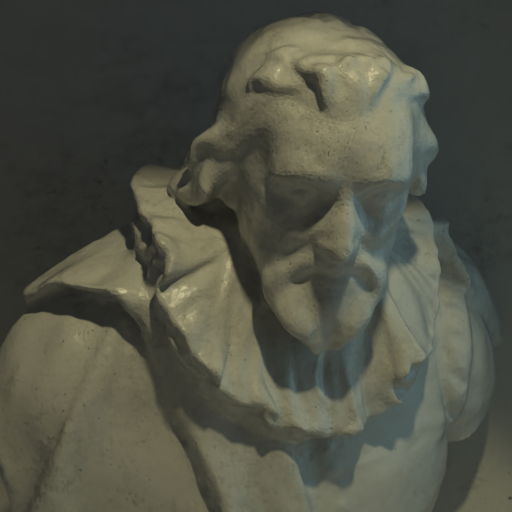} &
\includegraphics[width=0.2\linewidth]{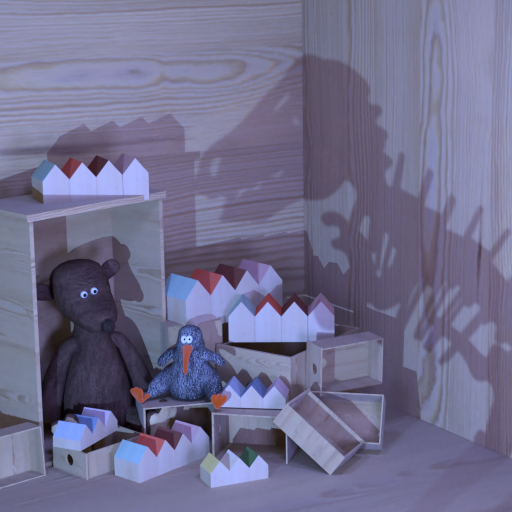} \\
Weight distribution & boxes(5,5) & sideboard(5,5) & cotton(5,5) & dino(5,5)
\end{tabular}
\caption{Weight distribution and views reconstructed using NeRF-rand (1st row) and NeRF-lr (2nd row), with several scenes.}
\label{fig:rand_lr}
\end{figure*}

\subsection{Ablation Study}
In this section, we evaluate different aspects of the proposed method.

\subsubsection{Random initialization vs Low-rank initialization}
\label{sec:random_init}
As aforementioned in Sec.~\ref{sec:distillation}, instead of the proposed NeRF initialization and finetuning steps, one could directly train a distilled NeRF with randomly initialized weights. Although this variant has a simpler training schedule, we show that it severely degrades the quality of the reconstruction. To evaluate the interest of the proposed distillation approach, we adopted the architecture of the distilled NeRF with rank $40$, and respectively initialized it using low-rank weights (noted as NeRF-lr) and random initialized weights drawn from Kaiming uniform distribution \cite{he2015delving} (noted as NeRF-rand). 
We then trained the two models until convergence. To guarantee the same experimental conditions, the two variants share the same camera pose and are trained without quantization.
Table~\ref{table:rand_lr} shows the PSNR of the light fields generated by NeRF-rand and NeRF-ft, with the same camera pose. The distilled NeRF trained with randomly initialized weights yields poor reconstruction quality. While NeRF trained with low-rank weights gives a good reconstruction quality. Fig.~\ref{fig:rand_lr} shows views reconstructed when using the two variants, and an example of weight distribution (for the light field `\textit{boxes}'). From the figure, we can observe that the weights of NeRF-rand and NeRF-lr follow different distributions. NeRF-lr has more values around 0 and long tails, NeRF-rand has fewer values around 0. Most weights are within the interval
$[-0.25,0.25]$ and the distribution does not have long tails. In terms of reconstruction
quality, NeRF-rand produces views with severe blurriness, while NeRF-lr gives views with subtle details, which means that it is better to use the low-rank NeRF to initialize the distilled version. The difference in terms of performance between NeRF-lr and NeRF-rand can be explained by the fact that, due to its low number of parameters of the distilled network, the optimization of the distilled network from scratch (with random initialization) may fall in local minima, giving a lower performance compared to the use of weights resulting from the low rank approximation of the large network.

\begin{table}[t]
\begin{center}
\caption{Average PSNR of views generated by NeRF initialized with random weights and with low-rank weights (with $r=40$). Both variants have fully converged.}
\label{table:rand_lr}
\scalebox{0.8}{
\begin{tabular}{c|c|c|c|c|c}
\hline
Variants & boxes & sideboard & dino & cotton & \textbf{average}\\
\hhline{------}
\hline
NeRF-rand & 25.18dB & 19.88dB & 23.27dB & 32.56dB & \textbf{25.22dB}\\
\hhline{------}
NeRF-lr & 37.85dB & 30.93dB & 40.90dB & 45.18dB & \textbf{38.72dB}\\
\hline
\end{tabular}}
\end{center}
\end{table}

\subsubsection{Quantization strategy analysis}
\label{sec:quant_stratgies}

\noindent
\textbf{Local quantization vs global quantization  }
Instead of using a global codebook of $N$ centroids for quantizing all layers, one can also utilize one codebook optimized for each layer. Table~\ref{table:quantization_option} shows the average PSNR and bitrate obtained when using the different quantization options: 
1). with 32-bit fixed-point scalar quantization, i.e. saving weights in 32-bit float type (DLR-NeRF-fpQuant);
2). with a quantizer optimized for each layer (DLR-NeRF-localQuant) and 3). a global codebook for all layers (DLR-NeRF-globalQuant). 
The local quantization suffers less from distortion but increases significantly the bitrate compared with the global quantization. While global quantization brings more distortion, its compression ratio is higher. 
Therefore, we adopted a global quantizer as it gives a lower bitrate with acceptable performance loss. Moreover, learning one codebook is much easier than learning several codebooks per layer.

\begin{table}[!t]
\begin{center}
\caption{Average PSNR and Bitrate obtained when using different quantization options: with fixed-point scalar quantization (DLR-NeRF-fpQuant), with one codebook for each layer (DLR-NeRF-localQuant), and when using a shared codebook for all layers (DLR-NeRF-globalQuant).}
\scalebox{0.9}{
\begin{tabular}{c|c|c}
\hline
Quantization options & PSNR & Bitrate\\
\hhline{---}
\hline
DLR-NeRF-fpQuant & 38.72dB & 0.30bpp\\
\hhline{---}
DLR-NeRF-localQuant & 38.70dB & 0.12bpp\\
\hhline{---}
DLR-NeRF-globalQuant & 38.18dB & 0.06bpp\\
\hhline{---}
\hline
\end{tabular}}
\end{center}
\label{table:quantization_option}
\end{table}

\subsubsection{Contribution of each block of the architecture}
To justify the contribution of each building block of the workflow, we measure the averaged PSNR and the model size after each step of our algorithm, and set original NeRF \cite{mildenhall2020nerf} (NeRF-org) as our baseline, with whose size being 100\%. The original NeRF has coarse and fine MLPs, and uses grund truth camera poses for synthesis. In our workflow, we instead adopt a simplified NeRF with only one MLP to learn scene information, and optimize it with a low rank constraint (LR-NeRF). The camera poses are estimated during the optimization of the LR-NeRF. Then this LR-NeRF is distilled into the DLR-NeRF with respect to the desired rank, here, we set desired the rank to $r=40$ for LR-NeRF and DLR-NeRF. DLR-NeRF is finally quantized with the number of centroids set to $N=256$ to obtain QDLR-NeRF.

Table~\ref{table:building_block} gives the PSNR and the model size evolution after each step of our workflow. The PSNR and size values are averaged over four synthetic scenes. We can observe from the table that, the transition from NeRF-org to LR-NeRF brings about 3dB PSNR loss. The loss is due to two factors:
1.) The camera poses are estimated instead of being ground truth camera poses, which introduces some errors 2.) The low rank constraint slightly decreases the synthesis performance. 
One could also think that this loss is caused by using only one MLP for synthesis. However, with most of light fields, the default sampling rate of $128$ along light rays is enough for one single MLP to yield high quality reconstruction. The replacement of two MLPs by one barely bring any performance loss.  

From LR-NeRF to DLR-NeRF, by distilling the network with the desired rank $r=40$, we reduce the model size by a factor of $3$ with only a 0.1dB PSNR loss. This small loss means, that the target rank of $40$ (the full rank being 256) can already store most of scene information for reconstruction, which also proves the effectiveness of our low rank approximation. 
From DLR-NeRF to QDLR-NeRF, the network is quantized from the first to the last layer, this gradual quantization step effectively reduces the model size to 1/5, with a limited performance loss of 0.6dB. 
In summary, thanks to the proposed low rank optimization, network distillation and quantization operations, we ultimately reduce the model size from 100\% (the reference being the bit number of the original NeRF) to 3.3\%, which is crucial in a context of light field compression, but also simply in terms of storage requirement for the NeRF model.
\begin{table}[t]
\begin{center}
\caption{PSNR averaged on the $8$ test light fields at the different steps of the proposed workflow: Original NERF (NeRF-org), NERF with the low rank constraint (LR-NeRF), Distilled low-rank NeRF (DLR-NeRF) and after quantization of DLR-NeRF (QDLR-NeRF).}
\scalebox{0.9}{
\begin{tabular}{c|c|c|c|c}
\hline
Metrics & NeRF-org & LR-NeRF & DLR-NeRF & QDLR-NeRF \\
\hhline{-----}
\hline
PSNR & 41.98dB & 38.83dB & 38.72dB & 38.18dB \\
\hhline{-----}
Size & 100\% & 50\% & 16.7\% & 3.3\% \\
\hline
\end{tabular}}
\end{center}
\label{table:building_block}
\end{table}

\section{Discussion}
Although our proposed method is able to effectively compress light fields, due to the fact that, it is based on the NeRF method, factors that limit the performance of NeRF will consequently become limitations of our method. For example, when applying NeRF to learn light fields captured by Lytro, blurriness and vignetting in each sub-aperture view may prevent the network from reconstructing precise scenes. As a result, the compression performance of our QDLR-NeRF is limited by these artifacts. While better performance is obtained on synthetic data, as they do not include these artifacts. That also explains why the gaps between the rate-distortion curves of our method and those of the reference methods on synthetic light fields are larger than those on Lytro-captured light fields.

Finally, please note that, even if we have chosen the original NeRF to demonstrate the interest of the proposed method, the method can obviously apply to different variants of NeRF implementations, e.g., SIREN\cite{sitzmann2019siren}, PlenOctrees\cite{Yu_2021_ICCV}, including to the recent SIGNET solution \cite{Feng_2021_ICCV}. 

\section{Conclusion}
In this paper, we have proposed a Quantized Distilled Low Rank Neural Radiance Field (QDLR-NeRF) representation for light field compression. The method optimizes a NeRF with a low rank constraint using an ADMM optimization framework to obtain the so-called LR-NeRF, the weights in LR-NeRF are low-rank, hence can be decomposed into a Tensor Train format. The decomposed weights are then used to initialize a distilled version of NeRF named DLR-NeRF, which has fewer parameters. Finally, we gradually quantize the layers of DLR-NeRF from the first layer to the last one using a global codebook learned by k-means, to get QDLR-NeRF. The distillation operation avoids tackling the rank-constrained optimization and the rate-constrained quantization simultaneously, making the entire pipeline work stably to compress a light field. The low rank optimization and quantization operations effectively reduce the size of the model without significantly degrading the reconstruction performance. Experimental results, using both synthetic and real-world light fields, show that our method outperforms the reference methods with a large margin. Furthermore, due to the fact that our method is based on synthesis method, it also has the advantage of generating novel views as densely as possible.

\bibliographystyle{model1-num-names}

\bibliography{biblio
}





\end{document}